\documentclass[twoside]{article}

\usepackage{microtype}
\usepackage{graphicx}
\usepackage{subfigure}
\usepackage{booktabs} 

\usepackage{hyperref}
\usepackage{float}

\usepackage{amsmath}
\usepackage{amssymb}
\usepackage{mathtools}
\usepackage{amsthm}

\usepackage[capitalize,noabbrev]{cleveref}

\theoremstyle{plain}
\newtheorem{theorem}{Theorem}[section]

\newtheorem{lemma}[theorem]{Lemma}

\theoremstyle{definition}
\newtheorem{definition}[theorem]{Definition}

\theoremstyle{remark}
\newtheorem{remark}[theorem]{Remark}

\usepackage[textsize=tiny]{todonotes}
\usepackage{amsmath}
\usepackage{mathtools}
\usepackage{amsfonts}
\usepackage{bm}
\usepackage{color}
\usepackage{xcolor}
\usepackage{amsthm}
\usepackage{booktabs}
\usepackage{enumitem}
\usepackage{multirow}
\usepackage{graphicx}
\usepackage{subfigure}
\usepackage{wrapfig}
\usepackage{siunitx}

\DeclareMathOperator*{\argmax}{arg max}
\DeclareMathOperator*{\argmin}{arg min}
\usepackage{mathtools}
\usepackage{adjustbox}

\DeclarePairedDelimiter\floor{\lfloor}{\rfloor}
\usepackage[accepted]{aistats2024}

\begin{document}

%

%

\twocolumn[

\aistatstitle{Pixel-wise Smoothing for Certified Robustness against Camera Motion Perturbations}
\vspace{-1.2mm}
\vphantom{\hphantom{\aistatsauthor{ Hanjiang Hu \And Zuxin Liu \And Linyi Li \And Jiacheng Zhu \And Ding Zhao}}}

\aistatsauthor{ Hanjiang Hu \And Zuxin Liu \And Linyi Li \vphantom{ whatever}}
\aistatsaddress{ Machine Learning Department,\\ Carnegie Mellon University \And  Mechanical Engineering,\\ Carnegie Mellon University  \And Computer Science, University of\\Illinois at Urbana-Champaign }
\aistatsauthor{  Jiacheng Zhu \And  Ding Zhao}
\aistatsaddress{ Computer Science and Artificial Intelligence Laboratory,\\Massachusetts Institute of Technology \And Mechanical Engineering,\\ Carnegie Mellon University  } ]

\begin{abstract}
Deep learning-based visual perception models lack robustness when faced with camera motion perturbations in practice. 
The current certification process for assessing robustness is costly and time-consuming due to the extensive number of image projections required for Monte Carlo sampling in the 3D camera motion space.
To address these challenges, we present a novel, efficient, and practical framework for certifying the robustness of 3D-2D projective transformations against camera motion perturbations. Our approach leverages a smoothing distribution over the 2D pixel space instead of in the 3D physical space,  eliminating the need for costly camera motion sampling and significantly enhancing the efficiency of robustness certifications. With the pixel-wise smoothed classifier, we are able to fully upper bound the projection errors  using a technique of uniform partitioning in camera motion space.
Additionally, we extend our certification framework to a more general scenario where only a single-frame point cloud is required in the projection oracle. 
Through extensive experimentation, we validate the trade-off between effectiveness and efficiency enabled by our proposed method. Remarkably, our approach achieves approximately 80\% certified accuracy while utilizing only 30\% of the projected image frames.
\end{abstract}

\section{INTRODUCTION}

Visual perception has been boosted in recent years by leveraging the power of neural networks, with broad applications in robotics and autonomous driving. Despite the success of deep learning based perception, robust perception suffers from  external sensing uncertainty in real-world settings, e.g. glaring illumination \cite{sun2022shift, hu2023seasondepth},  sensor placement perturbation \cite{hu2022investigating, li2023influence}, pose attack \cite{xu2022safebench},  motion blurring  and corruptions \cite{sayed2021improved, mintun2021interaction,kong2023robodepth}, etc. Besides, the internal vulnerability of deep neural networks has been well studied for years, and rich literature reveals that deep neural networks can be easily fooled by adversarial examples. The victim perception models predict incorrect results under stealthy perturbation on pixels \cite{goodfellow2014explaining,szegedy2013intriguing,xiao2018generating} 
or 2D semantic transformation, e.g. image rotation, scaling, translation, and other pixel-wise deformations of vector fields \cite{pei2017towards,dreossi2018semantic, hosseini2018semantic, engstrom2019exploring, hendrycks2018benchmarking, kanbak2018geometric, liu2018beyond}.

In response to such internal model vulnerability from $\ell_p$-bounded pixel-wise perturbations, in parallel to many empirical defense methods \cite{madry2017towards,tramer2018ensemble,ma2018characterizing, tramer2020adaptive}, lots of recent work provide provable robustness guarantees and certifications for any bounded perturbations within certain $\ell_p$ norm threshold \cite{cohen2019certified, tjeng2018evaluating, zhang2018efficient,dathathri2020enabling}. 
As for the challenging 2D semantic transformations including 2D geometric transformation, deterministic verification \cite{balunovic2019certifying, mohapatra2020towards, ruoss2021efficient, yang2022provable,hu2023robustness} and probabilistic certification methods \cite{fischer2020certified, li2021tss,  alfarra2021deformrs, hao2022gsmooth} are recently proposed to guarantee such robustness, which is more relevant and important to real-world applications.



\begin{figure*}
    \centering
    \includegraphics[width=0.95\textwidth]{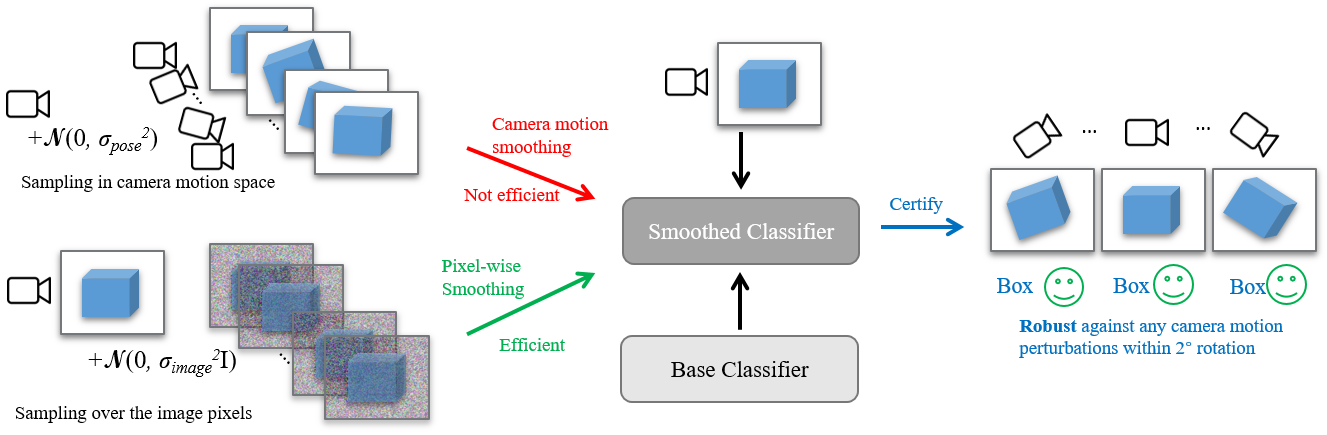}
    \vspace{-1mm}
    \caption{Overview of the robustness certification using pixel-wise smoothing (\textcolor{green}{green}) to avoid non-efficient  sampling in the camera motion space with too many projected frames required (\textcolor{red}{red}).}
    \label{fig:overview}
    \vspace{-3mm}
\end{figure*}

However, rare literature focuses on the more commonly-seen 3D-2D projective transformation caused by external sensing uncertainty of camera motion perturbations, which commonly exist in real-world applications such as autonomous driving and robotics. The external perturbations may cause severe consequences in safety-critical scenarios if the perception models are fooled by certain translation or rotation changes of the onboard cameras.
Recent work CMS \cite{hu2022robustness} studies how the camera motions influence the perception models and presents a resolvable robustness certification framework for such motion perturbations through randomized smoothing in the parameterized camera motion space. 
However,  although CMS gives the tight certification as an upper bound due to the formulation of the resolvable projective transformation \cite{li2021tss, hao2022gsmooth},   the high computational cost of "camera shaking" induced by the Monte Carlo sampling in the camera motion space  and the requirement of the entire dense point cloud of the object model restricts its practical applications. 

To this end, as shown in Figure \ref{fig:overview}, we introduce a new efficient  robustness certification framework with pixel-wise smoothing  against camera motion perturbations in a non-resolvable manner. We first construct smoothed classifiers over the pixel level, reducing the cost of redundant image projections in Monte Carlo sampling. Then we propose to use the uniform partitions technique with the consistent  camera motion interval  to fully cover the projection error based on the pixel-wise smoothed classifier. To further avoid the projection oracle where the whole dense point cloud must be required, we leverage the Lipschitz property of the projection function to approximate the upper bound of the partitioning intervals. This results in the successful certification given the projection oracle with knowing only one-frame point cloud, which is more convenient to obtain in many real-world applications. In addition to the theoretical contributions, we conduct extensive experiments to show the significant trade-off of required projection frames and certified accuracy  compared to the resolvable baseline as an upper bound, validating the efficiency and effectiveness of our proposed method. Our  contributions are summarized below.


\begin{itemize}[leftmargin=*]
\item We propose a new efficient robustness certification framework against camera motion perturbations using pixel-wise smoothing based on the uniform partitioning of  camera motion,  avoiding the substantial expenses associated with projection sampling in camera motion space.
\item We further derive a Lipschitz-based approximation for the upper bound of the partitioning interval and extend our theoretical framework to the general case with only the one-frame point cloud known in the projection oracle.
\item 
Comprehensive comparison experiments show that our method is much more efficient on the MetaRoom dataset: it requires only 30\% projected image frames to achieve 80\% certified accuracy compared to the upper bound of the resolvable baseline. 
\end{itemize}

\section{RELATED WORK}
\textbf{Provable Defenses against $\ell_p$-bounded Attacks}.
Compared to empirical defense approaches \cite{madry2017towards,tramer2018ensemble,ma2018characterizing, samangouei2018defense, tramer2020adaptive,pang2022robustness} to train robust models against specific adversarial perturbations, defense with provable guarantees aims to guarantee the accuracy for all perturbations within some $\ell_p$-bounded attack radius \cite{li2020sok,liu2019algorithms}. The complete certifications \cite{katz2017reluplex,ehlers2017formal} are NP-complete for deep neural networks \cite{li2020sok, zhang2022care}, though they guarantee finding existing attacks. Incomplete certifications are more practical as they provide relaxed guarantees to find non-certifiable examples, which can be categorized into deterministic and probabilistic ones \cite{ tjeng2018evaluating, wong2018provable,singh2019abstract,dathathri2020enabling,muller2022prima,zhang2022general}. Deterministic certifications adopt linear programming \cite{salman2019convex,zhang2018efficient} or semi-definite programming \cite{raghunathan2018certified,raghunathan2018semidefinite}, but they cannot scale up well to large datasets. Probabilistic certifications \cite{cohen2019certified} based on randomized smoothing show impressive scalability and promising advantages through adversarial training \cite{salman2019provably} and consistency regularization \cite{jeong2020consistency}. Lots of work show that the robustness certification can be boosted by improving robustness against Gaussian noise with some denoisers \cite{salman2020denoised,carlini2022certified}.


\textbf{Semantic Transformation Robustness Certification}.
The robustness of deep neural networks against real-world semantic transformations, e.g. translation, rotation, and scaling \cite{pei2017towards,dreossi2018semantic, hosseini2018semantic, engstrom2019exploring, hendrycks2018benchmarking, kanbak2018geometric, liu2018beyond}, presents challenges due to the complex optimization landscape involved in adversarial attacks.
Recent literature aims to provide the robustness guarantee against semantic transformations \cite{hao2022gsmooth,li2021tss, ruoss2021efficient, alfarra2021deformrs,balunovic2019certifying}, with either function relaxations-based deterministic guarantees \cite{balunovic2019certifying, mohapatra2020towards, lorenz2021robustness, ruoss2021efficient, yang2022provable} or random smoothing based high-confident probabilistic guarantees \cite{fischer2020certified, li2021tss, alfarra2021deformrs, chu2022tpc, hao2022gsmooth}. However, the robustness against projective transformation induced by sensor movement is rarely studied in the literature, while we believe it is a commonly seen perturbation source in practical applications. Recent  work \cite{hu2022robustness} proposes camera motion smoothing  to certify such robustness by leveraging smoothing distribution in the camera motion space. It is known to be tight as a resolvable transformation \cite{li2021tss, chu2022tpc, hao2022gsmooth}, which has the overlapped domain and support with smoothing distribution in the camera motion space. However, it requires high computational resources for Monte Carlo sampling with image projections as well as the expensive prior of the dense point cloud as an oracle. The above limitations motivate us to study a more efficient robustness certification method.

\begin{table*}[]
\centering
\caption{Table of notations for domain and function symbols}
\adjustbox{width=\textwidth}{
\begin{tabular}{ll|ll}
\hline
Domain Symbols & Meanings                           & Function Symbols & Meanings                              \\\hline
$\mathcal{X}\subset[0,1]^K\times\mathbb{Z}^2$              & 2D image with K channels           &   $\rho: \mathbb{P}\times\mathcal{Z}(\text{or }\mathcal{S})\rightarrow \mathbb{R}^2$               & 3D-2D position projection function    \\
$\mathcal{Z}\subset\mathbb{R}^6$              & Parameterized camera motion space  & $D: \mathbb{P}\times\mathcal{Z}(\text{or }\mathcal{S})\rightarrow \mathbb{R}$                 & Pixel-wise depth function             \\
$\mathcal{S}\subset\mathbb{R}\times\mathbf{0}^5$               & One-axis relative camera pose &    $\phi: \mathcal{X}\times\mathcal{Z}(\text{or }\mathcal{S})\rightarrow \mathcal{X}$               & 2D projective transformation function \\
$\mathbb{P}\subset\mathbb{R}^3$              & 3D points                          & $O: \mathcal{V}\times\mathcal{Z}(\text{or }\mathcal{S})\rightarrow \mathcal{X}$                   & 3D-2D pixel projection function       \\
$\mathcal{V}\subset\mathbb{R}^3 \times [0,1]^K$             & 3D points with K channels          & $p: \mathcal{Y}\mid\mathcal{X}\rightarrow \mathbb{R}$                & Score function in base classifier     \\
$\mathcal{Y}\subset\mathbb{Z}$             & Label space for classification     & $q: \mathcal{Y}\mid\mathcal{X};\varepsilon\rightarrow \mathbb{R}$                 & Score function in $\varepsilon$-smoothed classifier \\\hline
\end{tabular}
}
\end{table*}
\section{BACKGROUND AND PROBLEM FORMULATION}
\subsection{3D-2D Projective Transformation}
Following the literature in computer vision and graphics \cite{hartley2003multiple,shinya1991interference}, image projection can be obtained through the intrinsic matrix $\mathbf{K}$ of the camera and the extrinsic matrix of camera pose $\alpha=(\theta, t)\in\mathcal{Z}$ given dense 3D point cloud $\mathbb{P}\subset\mathbb{R}^3$ based on direct rasterization with z-buffering. In this way, each 3D point $P\in\mathbb{P}$ can be projected to a 2D position on the image plane through the 3D-2D projection function $\rho: \mathbb{P}\times\mathcal{Z}\to \mathbb{R}^2$ and pixel-wise depth function $D: \mathbb{P}\times\mathcal{Z}\rightarrow \mathbb{R}$. 
\begin{definition}[3D-2D position projection]
\label{def:oracle}
For any  3D point $P=(X, Y, Z)\in\mathbb{P}\subset\mathbb{R}^3$ under the camera coordinate with the camera intrinsic matrix $\mathbf{K}$, based on the camera motion of $\alpha= (\theta, t)  \in \mathcal{Z}\subset \mathbb{R}^6$ with  rotation matrix $ \mathbf{R} = \exp(\theta^\wedge) \in SO(3)$ and translation vector $t\in \mathbb{R}^3$, define the projection function $\rho: \mathbb{P}\times\mathcal{Z}\to \mathbb{R}^2$  and the depth function $D:\mathbb{P}\times\mathcal{Z}\rightarrow \mathbb{R}$  as
    \begin{align}
    \centering
    \label{def:projection}
    [\rho(P, \alpha), 1]^\top &= \frac{1}{D(P, \alpha)}\mathbf{K}\mathbf{R}^{-1}(P - t) \\
    D(P, \alpha) &= [0, 0, 1] \mathbf{R}^{-1}(P - t)
    \end{align}
\end{definition}
As defined in \cite{hu2022robustness}, given the $K$-channel colored  3D point cloud $V\in\mathcal{V}$, the 2D colored image $x\in\mathcal{X}$ can be obtained through 3D-2D projective transformation  $O: \mathcal{V}\times\mathcal{Z} \to \mathcal{X}$, as shown in Figure \ref{fig:projection}.
Based on this, 
define the 2D projective transformation  $\phi: \mathcal{X}\times\mathcal{Z}\to\mathcal{X}$ given the relative camera motion $\alpha\in\mathcal{Z}$. Specifically, if only one coordinate of $\alpha$ is non-zero, then it is denoted as one-axis relative camera motion $\alpha\in\mathcal{S}$.
\begin{definition}[3D-2D $K$-channel pixel-wise projection]
\label{def:proj_trans}
Given the position projection function $\rho: \mathbb{P}\times\mathcal{Z}\to \mathbb{R}^2$  and the depth function $D:\mathbb{P}\times\mathcal{Z}\rightarrow \mathbb{R}$ with $K$-channel 3D point cloud $V\in\mathcal{V}: \mathbb{P}\times[0, 1]^K$, for camera motion $\alpha\in\mathcal{Z}$, define the 3D-2D pixel projection function as $O: \mathcal{V}\times\mathcal{Z} \to \mathcal{X}$ to return $K$-channel image $x\in\mathcal{X}$,  $x=O(V, \alpha)$ where 
\begin{align}
        x_{k,r,s}&=  O(V, \alpha)_{k,r,s} =         V_{ P^*_\alpha, k} \\ \text{where }  P^*_\alpha &= \argmin \limits_{\{P\in \mathbb{P} \mid \floor{\rho(P,\alpha)} = (r,s) \}} D(P, \alpha)
\end{align}
where $\floor{\cdot}$ is the floor function. Specifically, if $x = O(V, 0)$, given the relative camera pose $\alpha\in\mathcal{Z}$, define the 2D projective transformation  $\phi: \mathcal{X}\times\mathcal{Z}\to\mathcal{X}$ as $\phi(x,\alpha)= O(V,\alpha)$.

\end{definition}
\subsection{Threat Model and Certification Goal}
We formulate the camera motion perturbation as an adversarial attack for the image classifier $g: \mathcal{X}\rightarrow\mathcal{Y}$, where there exists some camera pose $\alpha\in\mathcal{Z}$ under which the captured image $\phi(x,\alpha)$ fools the classifier $g$,   making a wrong prediction of object label. Specifically, if the whole dense point cloud $V$ is known as prior, we can obtain the image projection $\phi: \mathcal{X}\times\mathcal{Z}\to\mathcal{X}$ directly through 3D-2D projective transformation $O: \mathcal{V}\times\mathcal{Z} \to \mathcal{X}$, i.e.
 \begin{equation}
    \label{eq:entire_pro}
        x=\phi(x, 0) = O(V, 0), \phi(x,\alpha) = O(V, \alpha)
    \end{equation}

The certification goal is to provide robustness guarantees for vision models against all camera motion  perturbations within a certain radius in the  camera pose space through  3D-2D projective transformation. We formulate the goal as given the 2D projective transformation  $\phi$, finding a set $\mathcal{Z}_{\mathrm{adv}}\subseteq\mathcal{Z}_\phi$ for a classifier $g$ such that,
    \begin{equation}
    \label{eq:cert_goal}
        g(x) = g(\phi(x,\alpha)),  \forall \alpha\in\mathcal{Z}_{\mathrm{adv}}.
    \end{equation}

\section{PIXEL-WISE SMOOTHING CERTIFICATION METHOD}
\subsection{Pixel-wise Smoothed Classifier with  Projection}

To achieve the certification goal \eqref{eq:cert_goal}, we adopt the stochastic classifier based on the randomized smoothing \cite{cohen2019certified} over pixel level to construct the pixel-wise smoothed classifier, instead of the expensive smoothed classifier using smoothing distribution in camera motion space \cite{hu2022robustness}. Combining the image projection discussed above, we present the definition of smoothed classifier below, by adding zero-mean pixel-wise Gaussian noise to each projected image and taking the empirical mean as the smoothed prediction.
\begin{definition}[$\varepsilon$-smoothed classifier with 2D image projection]
\label{def:smooth_parametric_classifier}
Let $\phi: \mathcal{X}\times\mathcal{Z}\to\mathcal{X}$ be a 2D projective transformation
 parameterized with the one-axis relative camera motion $\alpha \in \mathcal{S}\subset \mathcal{Z}$ and $\varepsilon\in\mathcal{X}\sim\mathcal{P}_\varepsilon$  as the smoothing distribution. Let $x=\phi(x, 0)$ be under the original camera pose and $h: \mathcal{X}\to\mathcal{Y}$ be a base classifier $h(x) = \argmax_{y\in\mathcal{Y}} p(y\mid x)$. Under the one-axis relative camera pose $\alpha \in \mathcal{S}$, we define the $\varepsilon$-smoothed classifier $g:\mathcal{X}\to\mathcal{Y}$ as
 \begin{align}
     g(x;\alpha;\varepsilon) &= \argmax_{y\in\mathcal{Y}} q(y\mid \phi(x, \alpha);\varepsilon)\\  \text{where } q(y\mid \phi(x, \alpha);\varepsilon) &=  \mathbb{E}_{\varepsilon\sim\mathcal{P}_\varepsilon} p(y\mid \phi(x, \alpha) +\varepsilon)
 \end{align}
\end{definition}
\begin{remark}
    We remark that the definition of smoothed classifier \ref{def:smooth_parametric_classifier} is also applicable to the general 6-DoF translation and rotation of camera motion ($\alpha\in\mathcal{Z}$), but we mainly focus on the one-axis relative camera motion ($\alpha\in\mathcal{S}\subset\mathcal{Z}$) to make it consistent with the previous work \cite{hu2022robustness}.
\end{remark}
Based on the pixel-wise smoothing,  we then introduce the uniform partitions in camera motion space to fully cover all the pixel values within the camera motion perturbation radius. Specifically, we derive the upper bound of the partition interval  (PwS) and its Lipschitz-based approximation (PsW-L), which is further used to relax the prior from the entire dense point cloud to the one-frame point cloud in the projection oracle (PwS-OF).
\begin{figure*}
\centering
  \begin{minipage}[t]{0.32\linewidth}
    \centering
    \includegraphics[width=\linewidth]{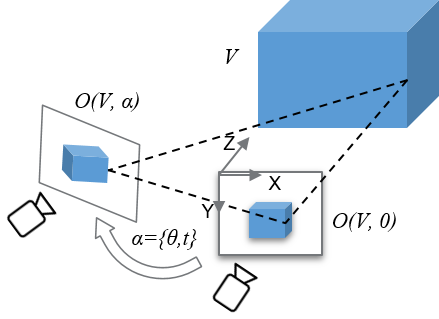}
    \caption{\small Image projection and coordinate framework in camera motion space.}
    \label{fig:projection}
  \end{minipage}%
  \hspace{1mm}
  \begin{minipage}[t]{0.32\linewidth}
    \centering
    \includegraphics[width=\linewidth]{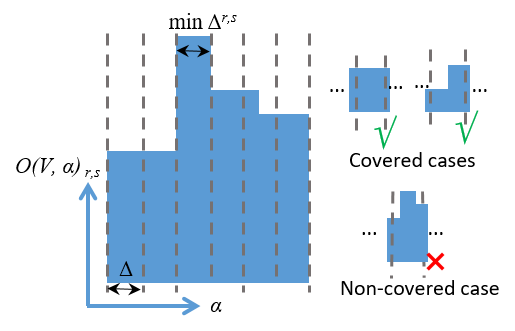}
    \caption{ \small Uniform partitions of $\Delta$ in camera motion $\alpha$ to fully cover all the pixel values $O(V,\alpha)_{r,s}$.}
    \label{fig:partitions}
  \end{minipage}
  \hspace{1mm}
  \begin{minipage}[t]{0.32\linewidth}
    \centering
    \includegraphics[width=\linewidth]{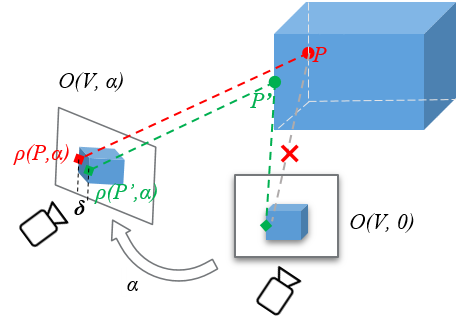}
    \caption{\small Projection from $P'$ as the one-frame point  and unknown $P$ in entire points  with $\delta$-convexity.}
    \label{fig:one-frame}
  \end{minipage}
\end{figure*}

\subsection{PwS: Certification via Pixel-wise Smoothing with Prior of  Entire Point Cloud}

Given the entire dense  colored point cloud $V: \mathbb{P}\times[0,1]^K$ for the image projection,  the projection function $O(V, \alpha)$ at each pixel $(r,s)\in\mathbb{Z}^2$ is a piecewise constant function w.r.t $\alpha$, as shown in Figure \ref{fig:partitions}. This is because the projected pixel value  is determined by  the target 3D point $P^*\in\mathbb{P}$ which has the least projected depth on the pixel $(r,s)$ for any camera pose within the motion interval $\mathbb{U}_{P^*,r,s}$, which is defined as \textit{consistent camera motion interval}, showing the set of views for which the same 3D scene point projects to a given pixel.

\begin{definition}[Consistent camera motion interval]
\label{def:consis_interval}
Given the 3D points $\mathbb{P}\subset\mathbb{R}^3$ and the projection function $\rho: \mathbb{P}\times\mathcal{Z}\to \mathbb{R}^2$  and the depth function $D: \mathbb{P}\times\mathcal{Z}\to \mathbb{R}$,  for any $P^*\in \mathbb{P}$ projected on $(r,s)$ with the least depth values, define the camera motion set $\mathbb{U}_{P^*,r,s}\subset\mathcal{Z}$ as the consistent camera motion interval,
\begin{align}
    \mathbb{U}_{P^*,r,s} = \{\alpha\mid P^*  = \argmin \limits_{\{P\in \mathbb{P} \mid \floor{\rho(P,\alpha)} = (r,s) \}} D(P, \alpha)\}
\end{align}
Specifically, for one-axis consistent camera motion interval $\mathbb{U}_{P,r,s}\subset\mathcal{S}$,  in the absence of ambiguity, we regard $\mathbb{U}_{P,r,s}$ as a subset of one-dimensional real number field based on non-zero coordinate in $\mathcal{S}$ in the following mathematical notations.
\end{definition}
Since  all the intervals of the piecewise constant function $O(V, \alpha)_{r,s}$ correspond to different 3D points projected to pixel $(r,s)$ as the camera motion $\alpha$ varies within one-axis camera motion perturbation $\mathcal{S}$, we introduce Lemma \ref{lemma:find_exact_Delta}, which demonstrates that for any given pixel $(r,s)$, an upper bound $\Delta^{r,s}$ exists for the \textit{consistent camera motion interval} $\mathbb{U}_{P,r,s}$, regardless of $P\in\mathbb{P}$. This upper bound ensures that all 3D-2D projections will fall within the projections of the endpoints of $\Delta^{r,s}$ for any camera motion that falls within $\Delta^{r,s}$. In this scenario, we describe the projection function $O(V, \alpha)_{r,s}$ as being \textit{fully covered} by this consistent interval upper bound $\Delta^{r,s}$, as illustrated in Figure \ref{fig:partitions}. The proof of Lemma \ref{lemma:find_exact_Delta} can be located in Appendix Section \ref{sec:B}.


\begin{lemma}[Upper bound of fully-covered motion interval]
\label{lemma:find_exact_Delta}
Given the  projection from entire 3D point  $V\in\mathcal{V}: \mathbb{P}\times[0, 1]^K$ along one-axis translation or rotation and the consistent camera motion interval $\mathbb{U}_{P,r,s}$  for any $P\in\mathbb{P}$ projected on $(r,s)$,
define the interval $\Delta^{r,s}$  as,
\begin{align}
    \Delta^{r,s} = \min_{P\in \mathbb{P} 
    }\{\sup{\mathbb{U}_{P,r,s}} - \inf{\mathbb{U}_{P,r,s}}\} 
\end{align}
 then for any projection on $(r,s)$ under camera motion $u\in\bigcup_{P\in\mathbb{P}}\mathbb{U}_{P,r,s}$, we have $\forall ~  0\leq\Delta^* \leq \Delta^{r,s}$
 \begin{align}
     O(V, u+\Delta^*)_{r,s} \in\{O(V, u)_{r,s},O(V, u+\Delta^{r,s})_{r,s}\} \notag
 \end{align}
\end{lemma}

Based on $\Delta^{r,s}$ in Lemma \ref{lemma:find_exact_Delta} for each pixel, we adopt uniform partitions over one-axis camera motion space $\mathcal{S}\subseteq \mathcal{Z}$, 
resulting in the robustness certification in Theorem \ref{theorem:diff_res}. The key to the proof is to upper bound the projection error \cite{li2021tss, chu2022tpc} in Equation \eqref{eq:errors_full}, which can be done through the fully-covered image partitions. The full proof of   Theorem \ref{theorem:diff_res} can be found in Appendix Section \ref{sec:B}.
\begin{theorem}[Certification with fully-covered partitions]
    \label{theorem:diff_res}
For the image projection from entire 3D point cloud $V\in\mathcal{V}: \mathbb{P}\times[0, 1]^K$, 
let $\phi$ be one-axis rotation or translation with parameters in $\mathcal{S}\subseteq\mathcal{Z}$  and uniformly partitioned $\{\alpha_i\}_{i=1}^N\subseteq\mathcal{S}$ with interval $\Delta_\alpha$, where
\begin{align}
    \Delta_\alpha \leq \min_{r,s}\Delta^{r,s} = \min_{r,s}\min_{P\in \mathbb{P}}\{\sup{\mathbb{U}_{P,r,s}} - \inf{\mathbb{U}_{P,r,s}}\} \notag
\end{align}
Let $y_A,y_B\in\mathcal{Y}$, $\varepsilon\sim\mathcal{N}(0, \sigma^2 \mathbb{I}_d)$ and suppose that for any $i$, the $\varepsilon$-smoothed classifier defined by $q(y\mid x;\varepsilon):=\mathbb{E}(p(y\mid x+\varepsilon))$ has  class probabilities that satisfy with top-2 classes $y_A,y_B$ as
                $ q(y_A\mid \phi(x, \alpha_i); \varepsilon) \geq p_A^{(i)} \geq p_B^{(i)} \geq \max_{y\neq y_A}q(y\mid \phi(x, \alpha_i); \varepsilon)$,
                
            then it is guaranteed that $\forall\alpha\in\mathcal{S}\colon y_A = \argmax_y q(y\mid \phi(x, \alpha);\varepsilon)$ if 
            \begin{equation}
            \label{eq:errors_full}
                \begin{gathered} \max_{1\leq i\leq N-1}\sqrt{\frac{1}{2} \sum_{k,r,s}(O(V, \alpha_i)_{krs}-O(V, \alpha_{i+1})_{krs})^2}    \\ < \frac{\sigma}{2}\min_{1\leq i \leq N}\left(\Phi^{-1}\left(p_A^{(i)}\right)-\Phi^{-1}\left(p_B^{(i)}\right)\right).
                \end{gathered}
            \end{equation}
\end{theorem}

\subsection{PwS-L: Certification via Pixel-wise Smoothing with Lipschitz-approximated Partitions}
\label{sec:lipschitz}

In more general cases without knowing the prior of the entire point cloud, the projected pixels within the \textit{consistent camera motion interval} $\mathbb{U}_{P, r, s}$ are easier to find  compared to $\mathbb{U}_{P, r, s}$ itself using Definition \ref{def:consis_interval}. In this case, 
we propose to approximate the upper bound of the fully-covered interval $\Delta^{r,s}$ as $\Delta_L^{r,s}$  by leveraging the Lipschitz property of projection oracle, as shown in Lemma \ref{lemma:find_Delta_appro}. 
\begin{lemma}[Approximated upper bound of fully-covered interval]
\label{lemma:find_Delta_appro}
For the  projection from entire 3D point  $V\in\mathcal{V}: \mathbb{P}\times[0, 1]^K$, given the one-axis monotonic position projection function $\rho$ in $l_\infty$ norm over camera motion space,
if the Lipschitz-based interval $\Delta_L^{r,s}$ is defined as
\begin{align}
    \centering
        \label{equ:Delta_L_appro}
   \Delta_L^{r,s}=\min\limits_{P\in \mathbb{P} 
   } \frac{1}{ L_{P}}
   \mid\rho(P,\sup{\mathbb{U}_{P,r,s}})-\rho(P,\inf{\mathbb{U}_{P,r,s}})\mid_\infty \notag
\end{align}
where 
$L_P$ is the  Lipschitz constant for projection function $\rho$ given 3D point $P$,
\begin{align}
    \centering
L_P = \max \{\max_{\alpha\in\mathcal{S}}|\frac{\partial\rho_1(P,\alpha)}{\partial\alpha}|, \max_{\alpha\in\mathcal{S}}|\frac{\partial\rho_2(P,\alpha)}{\partial\alpha}|\}
\end{align}
 then it holds that  $\Delta_L^{r,s} \leq \Delta^{r,s} = \min_{P\in \mathbb{P}}\{\sup{\mathbb{U}_{P,r,s}} - \inf{\mathbb{U}_{P,r,s}}\}  $.
    
\end{lemma}

 Based on the monotonicity of projection over camera motion and Theorem \ref{theorem:diff_res}, Theorem \ref{theorem:diff_res_appro} below  holds  for more general cases but with a smaller approximated fully-covered interval $\Delta_L^{r,s}$ and more uniform partitions $N$ as a trade-off  compared to Theorem \ref{theorem:diff_res}.
The full proof of Lemma \ref{lemma:find_Delta_appro} and Theorem \ref{theorem:diff_res_appro}  can be found in Appendix Section \ref{sec:C}.

\begin{theorem}[Certification with approximated partitions]
    \label{theorem:diff_res_appro}
For the projection from entire 3D point cloud $V\in\mathcal{V}: \mathbb{P}\times[0, 1]^K$ through the monotonic position projection function $\rho$ in $l_\infty$ norm over camera motion space, 
let $\phi$ be one-axis rotation or translation with parameters in $\mathcal{S}\subseteq\mathcal{Z}$  and uniformly partitioned $\{\alpha_i\}_{i=1}^N\subseteq\mathcal{S}$ with interval $\Delta_\alpha$, where
\begin{align}
    \centering
   \Delta_\alpha 
   &\leq \min \limits_{r,s} \min\limits_{P\in \mathbb{P} } \frac{1}{ L_{P}}
   |\rho(P,\sup{\mathbb{U}_{P,r,s}})  -\rho(P,\inf{\mathbb{U}_{P,r,s}})|_\infty \notag \\
 L_P &= \max \{\max_{\alpha\in\mathcal{S}}|\frac{\partial\rho_1(P,\alpha)}{\partial\alpha}|, \max_{\alpha\in\mathcal{S}}|\frac{\partial\rho_2(P,\alpha)}{\partial\alpha}|\}
\end{align}
Let $y_A,y_B\in\mathcal{Y}$, $\varepsilon\sim\mathcal{N}(0, \sigma^2 \mathbb{I}_d)$ and suppose that for any $i$, the $\varepsilon$-smoothed classifier defined by $q(y\mid x;\varepsilon):=\mathbb{E}(p(y\mid x+\varepsilon))$ has class probabilities that satisfy  with top-2 classes $y_A,y_B$ as
$
                 q(y_A\mid \phi(x, \alpha_i); \varepsilon) \geq p_A^{(i)} \geq p_B^{(i)} \geq \max_{y\neq y_A}q(y\mid \phi(x, \alpha_i); \varepsilon)
$,

            then it is guaranteed that $\forall\alpha\in\mathcal{S}\colon y_A = \argmax_y q(y\mid \phi(x, \alpha);\varepsilon)$ if 
            \begin{align}
                &\max_{1\leq i\leq N-1}\sqrt{\frac{1}{2} \sum_{k,r,s}(O(V, \alpha_i)_{krs}-O(V, \alpha_{i+1})_{krs})^2}  \notag\\<& \frac{\sigma}{2}\min_{1\leq i \leq N}\left(\Phi^{-1}\left(p_A^{(i)}\right)-\Phi^{-1}\left(p_B^{(i)}\right)\right)
            \end{align}
\end{theorem}

\subsection{PwS-OF: Certification via Pixel-wise Smoothing with One-Frame Point Cloud as Prior}

In this section, we extend our discussion to a more generalized scenario where only a one-frame dense point cloud is known. Such data can be conveniently acquired through various practical methods such as stereo vision \cite{gennery1977stereo}, depth cameras \cite{izadi2011kinectfusion}, or monocular depth estimation \cite{eigen2014depth}. Before delving into the Lipschitz properties of the projection oracle based on a single-frame point cloud, we begin by defining the image projection derived from the one-frame point cloud with $\delta$-convexity, showing how close  the pixels projected from entire 3D points are to the pixels projected from the  one-frame point cloud, as shown in Figure \ref{fig:one-frame}.
\begin{definition}[3D projection from one-frame point cloud with $\delta$-convexity]
\label{def:delta_conv}
Given a one-frame point cloud $\mathbb{P}_1\subset\mathbb{R}^3$, define the  $K$-channel image $x\in\mathcal{X}: [0,1]^K\times\mathbb{Z}^2$  as
\begin{align}
        x_{k,r,s}&= O(V, 0)_{k,r,s} =V_{P_1, k} \\ \text{ where } P_1 &\in  \mathbb{P}_1 \text{ s.t. }  \floor{\rho(P_1,\alpha)} = (r,s)
\end{align}
Define  $\delta$-convexity that with the minimal $\delta>0$,  for any new point $P \notin \mathbb{P}_1$ and any $\alpha\in\mathcal{S}$,   there exists $P' \in \mathbb{P}_1$ where
$\mid \rho(P, \alpha) - \rho(P', \alpha)\mid_\infty \leq \delta$, it holds that
$D(P, \alpha) \geq D(P', \alpha)$.
\end{definition}
Following the Lipschitz-based approximation of the upper bound of the fully-covered interval in Section \ref{sec:lipschitz},  we further approximate the fully-covered upper bound of the interval based on $\delta$-convexity point cloud in Lemma \ref{lemma:find_Delta_appro_one}, followed by the certification in Theorem \ref{theorem:diff_res_appro_one}. Note that the finite constant $C_\delta$ in Lemma \ref{lemma:find_Delta_appro_one} and Theorem \ref{theorem:diff_res_appro_one} can be expressed in the closed form 
in Appendix Lemma \ref{lemma:find_C_delta} together with the full proof in Appendix Section \ref{sec:D}.
\begin{lemma}[Approximated upper bound of interval from one-frame point cloud]
\label{lemma:find_Delta_appro_one}
Given the projection from one-frame 3D point cloud $V_1\in\mathcal{V}_1: \mathbb{P}_1\times[0, 1]^K$ and unknown entire point cloud $V\in\mathcal{V}: \mathbb{P}\times[0, 1]^K$ with $\delta$-convexity, if the one-axis position projection function $\rho$ is monotonic in $l_\infty$ norm over camera motion space, there exists a  finite constant $C_\delta\geq0$ such that 
if  the  interval $\Delta_{OF}^{r,s}$ is defined as,
\begin{align}
    \Delta_{OF}^{r,s} = \min \limits_{P_1\in\mathbb{P}_1}&\frac{1}{ L_{P_1} + C_\delta}\min_{P_1\in\mathbb{P}_1}( \mid\rho(P_1,\sup{\mathbb{U}_{P_1,r,s}}) \notag\\ &-\rho(P_1,\inf{\mathbb{U}_{P_1,r,s}})\mid_\infty-2\delta)
\end{align}
where 
$L_{P_1}$ is the  Lipschitz constant for projection function $\rho$ given 3D point $P_1$,
\begin{align}
    \centering
L_{P_1} = \max \{\max_{\alpha\in\mathcal{S}}|\frac{\partial\rho_1(P_1,\alpha)}{\partial\alpha}|, \max_{\alpha\in\mathcal{S}}|\frac{\partial\rho_2(P_1,\alpha)}{\partial\alpha}|\}
\end{align}
 then for any $P\in\mathbb{P}$, it holds that $ \Delta_{OF}^{r,s}  \leq \Delta^{r,s} = \min_{P\in \mathbb{P}}\{\sup{\mathbb{U}_{P,r,s}} - \inf{\mathbb{U}_{P,r,s}}\}$
\end{lemma}

\begin{theorem}[Certification with approximated partitions from one-frame point cloud]
    \label{theorem:diff_res_appro_one}
For the  projection from one-frame 3D point cloud $V_1\in\mathcal{V}_1: \mathbb{P}_1\times[0, 1]^K$ and unknown entire point cloud  $V\in\mathcal{V}: \mathbb{P}\times[0, 1]^K$ with $\delta$-convexity, 
let $\phi$ be the one-axis rotation or translation with parameters in $\mathcal{S}\subseteq\mathcal{Z}$ with  the  monotonic projection function $\rho$ in $l_\infty$ norm  over $\mathcal{S}$, we have uniformly partitioned $\{\alpha_i\}_{i=1}^N\subseteq\mathcal{S}$ under interval $\Delta_\alpha$ with  finite constant $C_\delta\geq0$ where
\begin{align}
    \Delta_\alpha \leq    \min_{r,s}\min \limits_{P_1\in\mathbb{P}_1}&\frac{1}{ L_{P_1} + C_\delta}  \min_{P_1\in\mathbb{P}_1}( \mid\rho(P_1,\sup{\mathbb{U}_{P_1,r,s}}) \notag \\&-\rho(P_1,\inf{\mathbb{U}_{P_1,r,s}})\mid_\infty-2\delta) \\
                 L_{P_1} = \max \{\max_{\alpha\in\mathcal{S}}&|\frac{\partial\rho_1(P_1,\alpha)}{\partial\alpha}|, \max_{\alpha\in\mathcal{S}}|\frac{\partial\rho_2(P_1,\alpha)}{\partial\alpha}|\}
\end{align}
Let $y_A, y_B\in\mathcal{Y}$, $\varepsilon\sim\mathcal{N}(0, \sigma^2 \mathbb{I}_d)$ and suppose that for any $i$, the $\varepsilon$-smoothed classifier defined by $q(y\mid x;\varepsilon):=\mathbb{E}(p(y\mid x+\varepsilon))$ has class probabilities that satisfy  with top-2 classes $y_A,y_B$ as
                 $q(y_A\mid \phi(x, \alpha_i); \varepsilon) \geq p_A^{(i)} \geq p_B^{(i)} \geq \max_{y\neq y_A}q(y\mid \phi(x, \alpha_i); \varepsilon)$,

            then it is guaranteed that $\forall\alpha\in\mathcal{S}\colon y_A = \argmax_y q(y\mid \phi(x, \alpha);\varepsilon)$ if 
            \begin{align}
                &\max_{1\leq i\leq N-1}\sqrt{\frac{1}{2} \sum_{k,r,s}(\phi(x, \alpha_i)_{krs}-\phi(x, \alpha_{i+1})_{krs})^2}  \notag\\ < &\frac{\sigma}{2}\min_{1\leq i \leq N}\left(\Phi^{-1}\left(p_A^{(i)}\right)-\Phi^{-1}\left(p_B^{(i)}\right)\right)
            \end{align}
\end{theorem}

\section{EXPERIMENTS}

In this section, we aim to answer two questions: the first one is can we  avoid too much "camera shaking" in the certification --- prevent sampling in camera motion space through the proposed method with much fewer projected frames required? The second question we want to answer is what is the trade-off  of the proposed method regarding efficiency and effectiveness compared to the resolvable baseline as an upper bound? We first get started with the experimental setup to answer these questions. The code is available at \url{https://github.com/HanjiangHu/pixel-wise-smoothing}. More details about experiments can be found in Appendix Section \ref{sec:E}.


\begin{table*}[htb]
\caption{Comparison of numbers of required projection frames and the percent ratio w.r.t. the resolvable baseline (100\%) for the same 10k Monte Carlo sampling.}
\centering
\begin{tabular}{ccccc}
\toprule
\begin{tabular}[c]{@{}c@{}}Num. of  Required \\ Projected Frames\end{tabular} 
 & CMS \cite{hu2022robustness}         &\begin{tabular}[c]{@{}c@{}}PwS \& \\ PwS-Diffusion\end{tabular}           & \begin{tabular}[c]{@{}c@{}}PwS-L \& \\ PwS-L-Diffusion\end{tabular}        & \begin{tabular}[c]{@{}c@{}}PwS-OF \& \\ PwS-OF-Diffusion\end{tabular}         \\\midrule
$T_z$, 10mm radius                 & 10k / 100\% & 3.5k / 35\%  & 5.1k / 51\% & 5.9k / 59\% \\
$R_y$, 0.25$^\circ$ radius                 & 10k / 100\% & 3.4k / 34\% & 5.0k / 50\% & 6.1k / 61\% \\\bottomrule
\end{tabular}
\label{tab:num_of_frames}
\end{table*}

\begin{table*}[h]
\caption{Certified accuracy and percent  w.r.t. the resolvable baseline as the upper bound (100\%)}
\centering
\begin{tabular}{ccccc}
\toprule
Radii along $T_z$ & CMS \cite{hu2022robustness}                    & PwS-Diffusion & PwS-L-Diffusion  & PwS-OF-Diffusion   \\ \midrule
10mm             & 0.491 / 100\%  & 0.392 / 79.8\% & 0.392 / 79.8\%  & 0.400 / 81.5\%  \\
20mm             & 0.475 / 100\% &  0.300 / 63.2\% & 0.292 / 61.5\%  & 0.300 / 63.2\%  \\ \bottomrule
\end{tabular}
\label{tab:certified_baseline}
\end{table*}

\begin{table}[htb]
\caption{Comparison of certified accuracy along z-axis translation and y-axis rotation }
\centering
\begin{tabular}{ccccc}
\toprule
\multirow{2}{*}{Projection type} & \multicolumn{2}{c}{Radii along $T_z$} & \multicolumn{2}{c}{Radii along $R_y$} \\
                              & 10mm       & 20mm      & 0.25$^\circ$       & 0.5$^\circ$       \\\midrule
PwS                           & 0.223      & 0.192     & 0.142      & 0.100     \\
PwS-L                         & 0.223      & 0.192     & 0.150      & 0.092     \\
PwS-OF                        & 0.231      & 0.192     & 0.150      &    0.108      \\ \bottomrule
\end{tabular}
\label{tab:certified_acc}
\end{table}

\begin{table*}[htb]
\caption{Certified accuracy of different smoothing variance $\sigma^2$ in all projections for PwS}
\centering
\begin{tabular}{ccccccc}
\toprule
Projection and radius & $T_z$, 10mm & $T_x$, 5mm & $T_y$, 5mm & $R_z$, 0.7$^\circ$ & $R_x$,  0.25$^\circ$ & $R_y$, 0.25$^\circ$ \\ \midrule
 $\sigma=0.25$    & 0.198   &   0.0      &   0.058      & 0.0        &      0.042     &  0.025        \\
$\sigma=0.5$    & 0.223   &   0.192      &    0.183     & 0.133        &   0.142        &    0.150      \\
$\sigma=0.75$     & 0.140    &  0.117       &  0.117       &     0.092    &     0.117      &  0.108  \\\bottomrule     
\end{tabular}
\label{tab:abla_variance}
\end{table*}

\begin{table*}[h]
\caption{Certified accuracy with different numbers of  partitioned images in pixel-wise smoothing}
\centering
\begin{tabular}{cccccccc}
\toprule
\begin{tabular}[c]{@{}c@{}}Number of Partitions\\ Radius 0.25$^\circ$,  y-axis rotation $R_y$\end{tabular} & 1000  & 2000  & 3000  & 4000  & 5000  & 6000  & 7000  \\ \midrule
ResNet50                                                                          & 0.083 & 0.117 & 0.142 & 0.142 & 0.150 & 0.150 & 0.150 \\
ResNet101                                                                         & 0.092 & 0.092 & 0.117 & 0.125 & 0.125 & 0.125 & 0.125 \\ \bottomrule
\end{tabular}
\label{tab:abla_partition}
\end{table*}

\begin{table}[ht]
\caption{Certified accuracy of different model complexity}
\centering
\begin{tabular}{ccccc}
\toprule Translation $T_z$                  & \multicolumn{2}{c}{10mm}                           & \multicolumn{2}{c}{100mm}                          \\
\multicolumn{1}{c}{ $\sigma$ of PwS  } & \multicolumn{1}{c}{$0.5$} & \multicolumn{1}{c}{$0.75$} & \multicolumn{1}{c}{$0.5$} & \multicolumn{1}{c}{$0.75$} \\\midrule
ResNet50                    & 0.223                   & 0.140                    & 0.142                   & 0.091                    \\
ResNet101                   & 0.150                   & 0.140                    & 0.108                   & 0.042   \\\bottomrule                  
\end{tabular}
\label{tab:abla_complexity}
\end{table}

\subsection{Experimental Setup}

\textbf{Dataset and  Smoothed Classifiers.} 
We adopt the MetaRoom dataset \cite{hu2022robustness} for the experiment, which contains camera poses associated with the entire dense point cloud. The dataset contains 20 different indoor objects for classification with camera motion perturbations of translation and rotation along x, y, and z axes ($T_z, T_x, T_z, R_z, R_x, R_y$). 
The default perception models are based on ResNet-50 and ResNet-101 is  used for different model complexity.
To enhance the robustness against  pixel-wise perturbation, the base classifiers are fine-tuned with both zero-mean Gaussian data augmentation \cite{cohen2019certified} and pre-trained diffusion-based denoiser \cite{carlini2022certified}, whose notation is with \textit{-Diffusion}. The default variance $\sigma^2$ of pixel-wise smoothing is with $\sigma=0.5$, while results for $\sigma=0.25, 0.75$ are also shown in the ablation study. With these  pixel-wise smoothed classifiers, the results of Theorem \ref{theorem:diff_res}, \ref{theorem:diff_res_appro}, \ref{theorem:diff_res_appro_one} are denoted as \textit{PwS}, \textit{PwS-L} and \textit{PwS-OF}, respectively. 
All the experiments are conducted on an Ubuntu 20.04 server with NVIDIA A6000 and 512G RAM.

\textbf{Baseline and Evaluation Metrics.}
We adopt the camera motion smoothing (CMS) method \cite{hu2022robustness} as the baseline on the MetaRoom dataset, which is based on the \textit{resolvable} projection with the tightest certification as an upper bound. 
The base classifiers are kept the same for fair comparisons.
Note that the diffusion-based denoiser \cite{carlini2022certified}, designed for pixel-wise Gaussian denoising, is not applicable to the CMS baseline whose input is without Gaussian noise.
We use the \textit{number of required projected frames} as a hardware-independent metric to evaluate certification efficiency, as image capture is the most resource-intensive part of certifying against camera motion. 
To assess certification effectiveness, we report \textit{certified accuracy} --  the ratio of the test images that are both correctly classified and satisfy the guarantee condition in the certification theorems \cite{li2021tss, chu2022tpc, hu2022robustness}, fairly showing how well the certification goal is achieved under the given  radius for different certification methods. 

\subsection{Efficient Certification Requiring Fewer Projected Frames}
We first show that our proposed certification is efficient with much fewer projected frames required. Frome Table \ref{tab:num_of_frames}, it can be seen that for the 10k Monte Carlo sampling to certify z-axis translation $T_z$ and y-axis rotation $R_y$, the baseline CMS \cite{hu2022robustness} needs 10k projected frames to construct the camera motion smoothed classifier, while our proposed certification only needs 30\% - 60\% projected frames as the partitioned images with pixel-wise smoothing of 10k Monte Carlo sampling. Besides, since the uniform partitioning can be done in the offline one-time manner, the proposed methods successfully avoid the impractical "camera shaking" dilemma in the robustness certification against camera motion.

We can also find that the certified accuracy of \textit{PwS}, \textit{PwS-L} and \textit{PwS-OF} are very close in Table \ref{tab:certified_acc} and \ref{tab:certified_baseline}, because the only difference is that they use different numbers of partitioned images as shown in Table \ref{tab:num_of_frames}. This is consistent with the theoretical analysis that our proposed three certification theorems  are theoretically supposed to have the same certified accuracy performance  but with different numbers of projected frames as partitions.

\begin{figure}[h]
    \centering
    \includegraphics[width=0.48\textwidth]{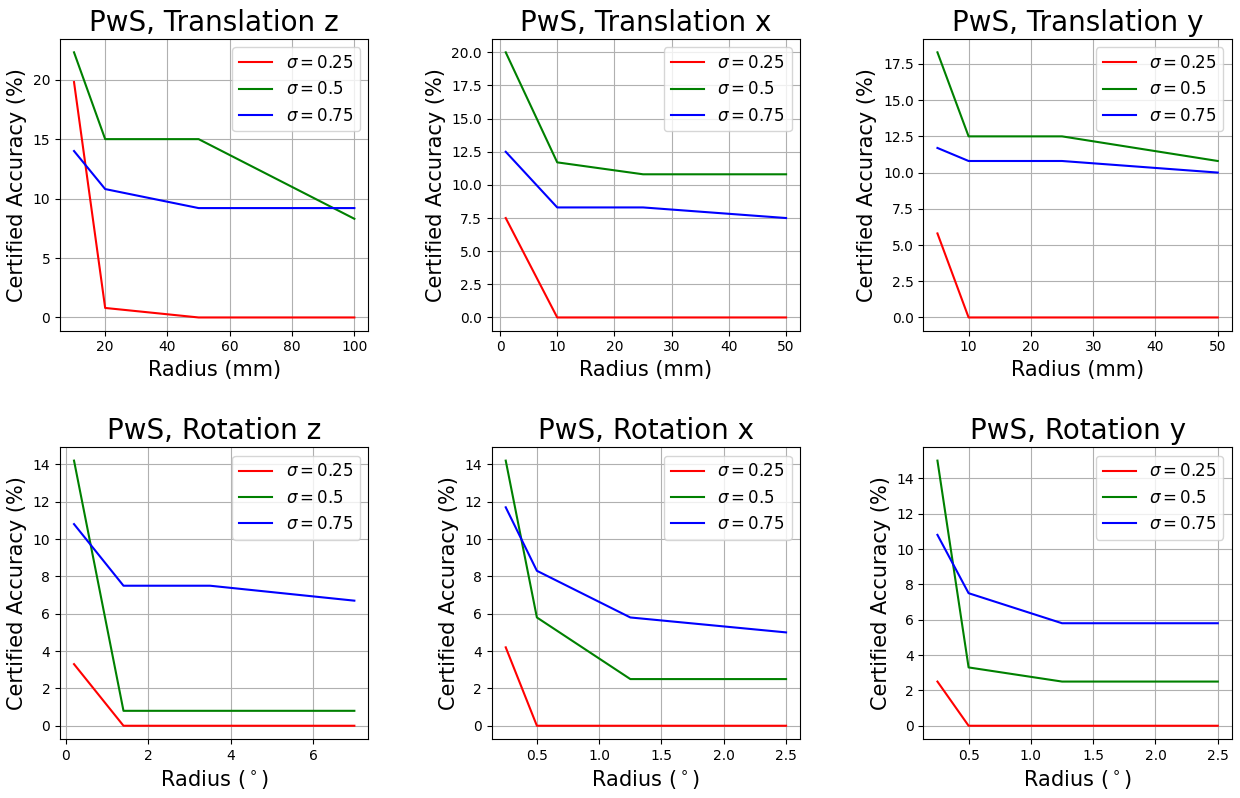}
    \caption{Certified accuracy of ResNet50 with smoothing variance $\sigma=0.25, 0.5, 0.75$ under different radii along $T_z, T_x, T_y, R_z, R_x, R_y$.}
    \label{fig:enter-label}
\end{figure}

\subsection{The trade-off of Certified Accuracy and Efficiency}
In this section, we present the trade-off regarding certified accuracy and the number of required projected frames compared to  the tightest certification CMS \cite{hu2022robustness}  as an upper bound as the resolvable case \cite{li2021tss,hao2022gsmooth}. In Table \ref{tab:certified_baseline},  although our proposed methods  with data augmentation are looser than CMS due to the pixel-wise smoothing with much fewer projected frames and better efficiency, the diffusion-based \textit{PwS} can remarkably boost the certified accuracy to achieve about 80\% of the upper bound with only 30\% of image projections in Table \ref{tab:num_of_frames}. Therefore, a significant trade-off can be seen by slightly sacrificing the certification effectiveness but saving a huge amount of image projection frames in certifications. 

\subsection{Ablation Study}
\label{sec:abla}
\textbf{Influence of the variance of the smoothing distribution.}
 As shown in Table \ref{tab:abla_variance}, for all translation and rotation axes,  the certified accuracy with $\sigma=0.5$ is  higher than  that with $\sigma=0.75$, showing the accuracy/robustness trade-off \cite{cohen2019certified}. However, the  performance with $\sigma=0.25$ is poor because the pixel-wise smoothing is too weak to cover the projection errors in the certification condition \eqref{eq:errors_full}.

 \textbf{Trend of certified accuracy along larger radii.}
 From Figure \ref{fig:enter-label}, it can be seen that given the smoothing variance, the certified accuracy generally decreases as the perturbation radii increase, showing that it is more likely to have non-robust adversarial samples in larger perturbation of camera motion. 
 However, the decay of certification is usually slower with larger radii, indicating that there exist some camera poses which are certifably robust against very large perturbation of camera motion. 
 Besides, under the smaller perturbation radii, $\sigma=0.5$ has the best performance, while certification with larger variance $\sigma=0.75$ performs better when the perturbation radius of rotation axes goes larger.

\textbf{Performance with different partitions.}
In this ablation, we discuss the influence of  partition numbers corresponding to the theory in empirical experiments under different models. As shown in Table \ref{tab:abla_partition}, it  can be seen that as the partition number goes up, the certification performance tends to converge, where the partitioned interval is  within the upper bound of the motion interval and is consistent with Lemma \ref{lemma:find_exact_Delta} and Table \ref{tab:num_of_frames}. 

\textbf{Performance under different model complexity.}
In Table \ref{tab:abla_complexity}, we compare the certified accuracy performance of different model complexity. We can find larger model complexity will be less certifiably robust with lower certified accuracy  under various variances and perturbation radii, which is consistent \cite{hu2022robustness} with previous work due to overfitting.

\section{CONCLUSION AND LIMITATIONS}
 In this work, we propose an efficient robustness certification framework against camera motion perturbation through pixel-wise smoothing. We introduce a new partitioning method to fully cover the projection errors through the pixel-wise smoothed classifier. Furthermore, we adopt the Lipschitz property of projection to approximate the partition intervals and extend our framework to the case of only requiring the one-frame point cloud. Extensive experiments show a significant trade-off  of using only 30\% projected frames but achieving 80\% certified accuracy compared to the upper bound baseline. 
 
 For the potential limitations,
our main theorems are mostly based on direct rasterization with z-buffering, which may differ from real-world imaging. However, we believe that our uniform partition method can also deal with the more complicated cases of existing interpolation errors based on \cite{li2021tss}. Besides, although we have relaxed the requirement of the entire point cloud in projection oracle in certification, we still need some prior point cloud and the assumption about the convexity of the scenes. Another limitation lies in that there is no real-world validation, especially in the setting of autonomous driving due to the lack of such outdoor dataset, although the MetaRoom dataset provides a realistic indoor environment, which also points out the future work for the community of computer vision and other applications. 
Regarding negative social impact, improved robustness in visual perception models might lead to increased surveillance capabilities and potential misuse, infringing on individual privacy and raising ethical concerns.

\clearpage
\subsubsection*{Acknowledgements}
We thank Dr. Bo Li from the University of Chicago for the insightful discussion and feedback.

{
\bibliographystyle{apalike}
\bibliography{ref}
}
\onecolumn
\appendix



\section{Preliminary Definitions and Theorems}
\label{sec:A}
\subsection{Preliminary Definitions}
\begin{definition}[3D-2D position projection, \textbf{restated} of Definition \ref{def:oracle} and Definition 1 from \cite{hu2022robustness}]
\label{def:oracle_re}
For any  3D point $P=(X, Y, Z)\in\mathbb{P}\subset\mathbb{R}^3$ under the camera coordinate with the camera intrinsic matrix $\mathbf{K}$, based on the camera motion of $\alpha= (\theta, t)  \in \mathcal{Z}\subset \mathbb{R}^6$ with  rotation matrix $ \mathbf{R} = \exp(\theta^\wedge) \in SO(3)$ and translation vector $t\in \mathbb{R}^3$, define the projection function $\rho: \mathbb{P}\times\mathcal{Z}\to \mathbb{R}^2$  and the depth function $D:\mathbb{P}\times\mathcal{Z}\rightarrow \mathbb{R}$  as
    \begin{align}
    \centering
    \label{def:projection_re}
    [\rho(P, \alpha), 1]^\top = \frac{1}{D(P, \alpha)}\mathbf{K}\mathbf{R}^{-1}(P - t), \quad 
    D(P, \alpha) = [0, 0, 1] \mathbf{R}^{-1}(P - t)
    \end{align}
\end{definition}
\begin{definition}[3D-2D $K$-channel pixel-wise projection, \textbf{restated} of Definition \ref{def:proj_trans} and Definition 2 from \cite{hu2022robustness}]
\label{def:proj_trans_re}
Given the position projection function $\rho: \mathbb{P}\times\mathcal{Z}\to \mathbb{R}^2$  and the depth function $D:\mathbb{P}\times\mathcal{Z}\rightarrow \mathbb{R}$ with $K$-channel 3D point cloud $V\in\mathcal{V}: \mathbb{P}\times[0, 1]^K$, for camera motion $\alpha\in\mathcal{Z}$, define the 3D-2D pixel projection function as $O: \mathcal{V}\times\mathcal{Z} \to \mathcal{X}$ to return $K$-channel image $x\in\mathcal{X}$,  $x=O(V, \alpha)$ where 
\begin{align}
        x_{k,r,s}=  O(V, \alpha)_{k,r,s} =         V_{ P^*_\alpha, k}, \quad \text{where } P^*_\alpha = \argmin \limits_{\{P\in \mathbb{P} \mid \floor{\rho(P,\alpha)} = (r,s) \}} D(P, \alpha)
\end{align}
where $\floor{\cdot}$ is the floor function. Specifically, if $x = O(V, 0)$, given the relative camera pose $\alpha\in\mathcal{Z}$, define the 2D projective transformation  $\phi: \mathcal{X}\times\mathcal{Z}\to\mathcal{X}$ as $\phi(x,\alpha)= O(V,\alpha)$.
\end{definition}

\begin{definition}[$\varepsilon$-smoothed classifier with 2D image projection, \textbf{restated} of Definition \ref{def:smooth_parametric_classifier}]
\label{def:smooth_parametric_classifier_re}
Let $\phi: \mathcal{X}\times\mathcal{Z}\to\mathcal{X}$ be a 2D projective transformation
 parameterized with the one-axis relative camera motion $\alpha \in \mathcal{S}\subset \mathcal{Z}$ and $\varepsilon\in\mathcal{X}\sim\mathcal{P}_\varepsilon$  as the smoothing distribution. Let $x=\phi(x, 0)$ be under the original camera pose and $h: \mathcal{X}\to\mathcal{Y}$ be a base classifier $h(x) = \argmax_{y\in\mathcal{Y}} p(y\mid x)$. Under the one-axis relative camera pose $\alpha \in \mathcal{S}$, we define the $\varepsilon$-smoothed classifier $g:\mathcal{X}\to\mathcal{Y}$ as
 \begin{align}
     g(x;\alpha;\varepsilon) = \argmax_{y\in\mathcal{Y}} q(y\mid \phi(x, \alpha);\varepsilon),  \text{where } q(y\mid \phi(x, \alpha);\varepsilon) =  \mathbb{E}_{\varepsilon\sim\mathcal{P}_\varepsilon} p(y\mid \phi(x, \alpha) +\varepsilon)
 \end{align}
\end{definition}
\begin{remark}
    We remark that the definition of the smoothed classifier \ref{def:smooth_parametric_classifier_re} is also applicable to the general 6-DoF translation and rotation of camera motion ($\alpha\in\mathcal{Z}$), but we mainly focus on the one-axis relative camera motion ($\alpha\in\mathcal{S}\subset\mathcal{Z}$) to make it consistent with the previous work \cite{hu2022robustness}.
\end{remark}

\subsection{Main Theorems for Transformation Specific Smoothing}
\begin{definition} (Differentially resolvable 3D-2D projective transformation)
\label{def:diff_resolvable_2d_proj}
            Let $\phi: \mathcal{X}\times\mathcal{Z}\to\mathcal{X}$ be a 2D projective transformation based on a 3D oracle $O: \mathcal{V}\times\mathcal{Z} \to \mathcal{X}$ with noise space $\mathcal{Z}_\phi$ and let $\psi:\mathcal{X}\times\mathcal{Z}_\psi\to\mathcal{X}$ be a resolvable 2D transformation with noise space $\mathcal{Z}_\psi$. We say that $\phi$ can be differentially resolved by $\psi$ if for any $V\in\mathcal{V}$ there exists function $\delta_V:\mathcal{Z}_\phi\times \mathcal{Z}_\phi \to \mathcal{Z}_\psi$ such that for any $\alpha \in \mathcal{Z}_\phi$ and any $\beta\in\mathcal{Z}_\phi$,
            \begin{equation}
                \phi(O(V, 0), \alpha) = \psi(\phi(O(V, 0), \beta),\delta_V(\alpha,\beta)).
            \end{equation}
            Specifically, the projective transformation from a discrete oracle can be resolved by the additive transformation by $$\psi(x,\delta) = x + \delta, \delta_V(\alpha,\beta) = \phi(O(V, 0), \alpha) - \phi(O(V, 0), \beta)$$
\end{definition}
Following the category of resolvable and non-resolvable  or differentially resolvable transformation in literature \cite{li2021tss, chu2022tpc, hao2022gsmooth}, we define the differentially resolvable property of 3D-2D projective transformation in Definition \ref{def:diff_resolvable_2d_proj}. Therefore, the following theorems are applicable as the foundation of other theorems in this paper.
\begin{theorem}[Theorem 2 in \cite{li2021tss}]
\label{thm:tss_dr}
            Let $\phi: \mathcal{X}\times\mathcal{Z}\to\mathcal{X}$ be a transformation that is resolved by $\psi:\mathcal{X}\times\mathcal{Z}_\psi\to\mathcal{X}$. Let $\varepsilon\sim \mathcal{P}_\varepsilon$ be a $\mathcal{Z}_\psi$-valued random variable and suppose that the smoothed classifier $g: \mathcal{X} \to \mathcal{Y}$ given by $q(y\mid x;\varepsilon) = \mathbb{E}(p(y\mid \psi(x, \varepsilon)))$ predicts $g(x; \varepsilon) = y_A = \argmax_y q(y\mid x;\varepsilon)$.
            Let $\mathcal{S}\subseteq\mathcal{Z}_\phi$ and $\{\alpha_i\}_{i=1}^N\subseteq\mathcal{Z}$ be a set of transformation parameters such that for any $i$, the class probabilities satisfy
            \begin{equation}
                q(y_A\mid \phi(x, \alpha_i); \varepsilon) \geq p_A^{(i)}\geq p_B^{(i)} \geq \max_{y\neq y_A} q(y\mid \phi(x, \alpha_i); \varepsilon).
            \end{equation}
            Then there exists a set $\Delta^*\subseteq\mathcal{Z}_\psi$ with the property that, if for any $\alpha\in\mathcal{S}, \exists\alpha_i$ with $\delta_x(\alpha, \alpha_i)\in\Delta^*$, then it is guaranteed that
            \begin{equation}
                q(y_A\mid \phi(x, \alpha);\varepsilon) > \max_{y\neq y_A}q(y\mid \phi(x, \alpha);\varepsilon).
            \end{equation}
\end{theorem}
The rigorous proof of Theorem \ref{thm:tss_dr} can be found in \cite{li2021tss}.
\begin{theorem}[Theorem 7 in \cite{chu2022tpc}, Corollary 2 in \cite{li2021tss}]
\label{lemma: diff_res}
Let $\psi(x, \delta) = x + \delta$ and let $\varepsilon\sim\mathcal{N}(0, \sigma^2 \mathbb{I}_d)$. Furthermore, let $\phi$ be a transformation with parameters in $\mathcal{Z}_\phi\subseteq\mathbb{R}^m$ and let $\mathcal{S}\subseteq\mathcal{Z}_\phi$ and $\{\alpha_i\}_{i=1}^N\subseteq\mathcal{S}$. Let $y_A\in\mathcal{Y}$ and suppose that for any $i$, the $\varepsilon$-smoothed classifier defined by $q(y\mid x;\varepsilon):=\mathbb{E}(p(y\mid x+\varepsilon))$ has class probabilities that satisfy
            \begin{equation}
                 q(y_A\mid \phi(x, \alpha_i); \varepsilon) \geq p_A^{(i)} \geq p_B^{(i)} \geq \max_{y\neq y_A}q(y\mid \phi(x, \alpha_i); \varepsilon).
             \end{equation}
            Then it is guaranteed that $\forall\alpha\in\mathcal{S}\colon y_A = \argmax_y q(y\mid \phi(x, \alpha);\varepsilon)$ if the maximum  projective error
            \begin{equation}
                \label{eq:max_interpolation_error}
                M_{\mathcal{S}}:=\max_{\alpha\in\mathcal{S}}\min_{1\leq i \leq N}\left\|\phi(x, \alpha) - \phi(x, \alpha_i)\right\|_2
            \end{equation}
            \begin{equation}
                \label{eq:general_Ms}
                \begin{gathered}
                    \text{satisfies}\quad
                    M_{\mathcal{S}} < R:=\frac{\sigma}{2}\min_{1\leq i \leq N}\left(\Phi^{-1}\left(p_A^{(i)}\right)-\Phi^{-1}\left(p_B^{(i)}\right)\right).
                \end{gathered}
            \end{equation}
\end{theorem}
The Theorem \ref{lemma: diff_res} is directly obtained if the projective transformation can be resolved by  additive transformation, as shown in Definition \ref{def:diff_resolvable_2d_proj}. We direct readers to \cite{li2021tss} for rigorous proof of Theorem \ref{lemma: diff_res}.

\section{Proofs in Certification via Pixel-wise Smoothing with Prior of  Entire Point Cloud (PwS)}
\label{sec:B}
Here we present the proof of Lemma \ref{lemma:find_exact_Delta} and Theorem \ref{theorem:diff_res}, which gives the certification using the fully-covered interval of uniform partitions and serves as the foundation of the following theorems. 
\begin{definition}[Consistent camera motion interval, \textbf{restated} of Definition \ref{def:consis_interval}]
\label{def:consis_interval_re}
Given the 3D points $\mathbb{P}\subset\mathbb{R}^3$ and the projection function $\rho: \mathbb{P}\times\mathcal{Z}\to \mathbb{R}^2$  and the depth function $D: \mathbb{P}\times\mathcal{Z}\to \mathbb{R}$,  for any $P^*\in \mathbb{P}$ projected on $(r,s)$ with the least depth values, define the camera motion set $\mathbb{U}_{P^*,r,s}\subset\mathcal{Z}$ as the consistent camera motion interval,
\begin{align}
    \mathbb{U}_{P^*,r,s} = \{\alpha\mid P^*  = \argmin \limits_{\{P\in \mathbb{P} \mid \floor{\rho(P,\alpha)} = (r,s) \}} D(P, \alpha)\}
\end{align}
Specifically, for one-axis consistent camera motion interval $\mathbb{U}_{P,r,s}\subset\mathcal{S}$,  in the absence of ambiguity, we regard $\mathbb{U}_{P,r,s}$ as a subset of one-dimensional real number field based on non-zero coordinate in $\mathcal{S}$ in the following mathematical notations.
\end{definition}

\subsection{Proof on Lemma \ref{lemma:find_exact_Delta}: Upper bound of fully-covered motion interval}
\begin{lemma}[Upper bound of fully-covered motion interval, \textbf{restated} of Lemma \ref{lemma:find_exact_Delta}]
\label{lemma:find_exact_Delta_re}
Given the  projection from entire 3D point  $V\in\mathcal{V}: \mathbb{P}\times[0, 1]^K$ along one-axis translation or rotation and the consistent camera motion interval $\mathbb{U}_{P,r,s}$  for any $P\in\mathbb{P}$ projected on $(r,s)$,
define the interval $\Delta^{r,s}$  as,
\begin{align}
    \Delta^{r,s} = \min_{P\in \mathbb{P} 
    }\{\sup{\mathbb{U}_{P,r,s}} - \inf{\mathbb{U}_{P,r,s}}\} 
\end{align}
 then for any projection on $(r,s)$ under camera motion $u\in\bigcup_{P\in\mathbb{P}}\mathbb{U}_{P,r,s}$, we have $\forall ~  0\leq\Delta^* \leq \Delta^{r,s}$
 \begin{align}
     O(V, u+\Delta^*)_{r,s} \in\{O(V, u)_{r,s},O(V, u+\Delta^{r,s})_{r,s}\}
 \end{align}
\end{lemma}
\begin{proof}
Considering the projection function $\rho$ on $r,s$ with any  3D point $P:\floor{\rho(P,\alpha)}=(r,s)$ at camera motion $\alpha\in\mathcal{S}$ with the least depth, $\mathbb{U}_{P,r,s}\neq \emptyset$. With
\begin{align}
\label{controdiction_re}
    \Delta^{r,s} = \min_{P\in \mathbb{P} 
    }\{\sup{\mathbb{U}_{P,r,s}} - \inf{\mathbb{U}_{P,r,s}}\}= \min_{P\in \mathbb{P} \mid \floor{\rho(P,\alpha)} = (r,s), \alpha\in\mathbb{U}_{P,r,s}}\sup{\mathbb{U}_{P,r,s}} - \inf{\mathbb{U}_{P,r,s}} 
\end{align}

For the 3D projective oracle $O$ on pixel $(r,s)$, based on the definition of $O(V, u), O(V, u+\Delta^{r,s} )$, we have $O(V, u)_{r,s} =         V_{P_u}, O(V, u+\Delta^{r,s} )_{r,s} =         V_{P_{u+\Delta^{r,s} }}$, where
$$ P_u = \argmin \limits_{\{P\in \mathbb{P} \mid \floor{\rho(P,u)} = (r,s) \}} D(P, u), P_{u+\Delta^{r,s} } = \argmin \limits_{\{P\in \mathbb{P} \mid \floor{\rho(P,u+\Delta^{r,s})} = (r,s) \}} D(P, u+\Delta^{r,s} )$$
Suppose there exists $\Delta^*\in[0,\Delta^{r,s}]$ such that $$O(V, u+\Delta^*)_{r,s} \neq O(V, u)_{r,s}, O(V, u+\Delta^{r,s})_{r,s} \neq O(V, u+\Delta^{r,s})_{r,s}$$
i.e.,  there exists $O(V, u+\Delta^*)_{r,s} = V_{P_{u+\Delta^*}}$ such that $P_{u} \neq P_{u+\Delta^*}, P_{u+\Delta^{r,s}}\neq P_{u+\Delta^*}$. In this case, according to the definition of $\mathbb{U}_{P_{u+\Delta^*},r,s}$, it holds that $$\sup{\mathbb{U}_{P_{u+\Delta^*},r,s}} - \inf{\mathbb{U}_{P_{u+\Delta^*},r,s}} < \Delta^{r,s}$$
which contradicts with \eqref{controdiction_re}.
Therefore,  such $O(V, u+\Delta^*)_{r,s}$ does not exist and for any $0\leq\Delta^*\leq \Delta^{r,s}$,
$$O(V, u+\Delta^*)_{r,s} \in\{O(V, u)_{r,s},O(V, u+\Delta^{r,s})_{r,s}\}$$
which concludes the proof.
\end{proof}

\subsection{Proof on Theorem \ref{theorem:diff_res}: Certification with fully-covered partitions}
\begin{theorem}[Certification with fully-covered partitions, \textbf{restated} of Theorem \ref{theorem:diff_res}]
    \label{theorem:diff_res_re}
For the image projection from entire 3D point cloud $V\in\mathcal{V}: \mathbb{P}\times[0, 1]^K$, 
let $\phi$ be one-axis rotation or translation with parameters in $\mathcal{S}\subseteq\mathcal{Z}$  and uniformly partitioned $\{\alpha_i\}_{i=1}^N\subseteq\mathcal{S}$ with interval $\Delta_\alpha$, where
\begin{align}
    \Delta_\alpha \leq \min_{r,s}\Delta^{r,s} = \min_{r,s}\min_{P\in \mathbb{P}}\{\sup{\mathbb{U}_{P,r,s}} - \inf{\mathbb{U}_{P,r,s}}\} \notag
\end{align}
Let $y_A,y_B\in\mathcal{Y}$, $\varepsilon\sim\mathcal{N}(0, \sigma^2 \mathbb{I}_d)$ and suppose that for any $i$, the $\varepsilon$-smoothed classifier defined by $q(y\mid x;\varepsilon):=\mathbb{E}(p(y\mid x+\varepsilon))$ has  class probabilities that satisfy with top-2 classes $y_A,y_B$ as
                $ q(y_A\mid \phi(x, \alpha_i); \varepsilon) \geq p_A^{(i)} \geq p_B^{(i)} \geq \max_{y\neq y_A}q(y\mid \phi(x, \alpha_i); \varepsilon)$,
                
            then it is guaranteed that $\forall\alpha\in\mathcal{S}\colon y_A = \argmax_y q(y\mid \phi(x, \alpha);\varepsilon)$ if 
            \begin{equation}
            \label{eq:errors_full_re}
                \begin{gathered} \max_{1\leq i\leq N-1}\sqrt{\frac{1}{2} \sum_{k,r,s}(O(V, \alpha_i)_{krs}-O(V, \alpha_{i+1})_{krs})^2}   < \frac{\sigma}{2}\min_{1\leq i \leq N}\left(\Phi^{-1}\left(p_A^{(i)}\right)-\Phi^{-1}\left(p_B^{(i)}\right)\right).
                \end{gathered}
            \end{equation}
\end{theorem}
\begin{proof}
According to Theorem \ref{lemma: diff_res}, we need to upper bound  $$\max_{\alpha\in\mathcal{S}}\min_{1\leq i \leq N}\left\|\phi(x, \alpha) - \phi(x, \alpha_i)\right\|_2$$ with uniform partitions under interval of $ \Delta_\alpha = \alpha_{i+1} - \alpha_{i}$. 
\begin{align}
    \centering
    &\max_{\alpha\in\mathcal{S}}\min_{1\leq i \leq N}\left\|\phi(x, \alpha) - \phi(x, \alpha_i)\right\|_2  
    =\max_{\alpha\in\mathcal{S}}\min_{1\leq i \leq N}\left\|O(V, \alpha) - O(V, \alpha_i)\right\|_2 \notag\\\leq & \max_{1\leq i\leq N-1}\sqrt{\max_{\alpha_i\leq \alpha \leq \alpha_{i+1}}\min\{(O(V, \alpha)-O(V, \alpha_i))^2, (O(V, \alpha)-O(V, \alpha_{i+1}))^2\} }\notag \\
    = & \max_{1\leq i\leq N-1}\sqrt{\max_{\alpha_i\leq \alpha \leq \alpha_{i+1}}\min\{\sum_{k,r,s}(O(V, \alpha)_{krs}-O(V, \alpha_i)_{krs})^2,  \sum_{k,r,s}(O(V, \alpha)_{krs}-O(V, \alpha_{i+1})_{krs})^2\}} \notag \\
    \leq & \max_{1\leq i\leq N-1}\sqrt{\max_{\alpha_i\leq \alpha \leq \alpha_{i+1}}\frac{1}{2}\{\sum_{k,r,s}(O(V, \alpha)_{krs}-O(V, \alpha_i)_{krs})^2+ \sum_{k,r,s}(O(V, \alpha)_{krs}-O(V, \alpha_{i+1})_{krs})^2\}} \notag \\
    = & \max_{1\leq i\leq N-1}\sqrt{\max_{\alpha_i\leq \alpha \leq \alpha_{i+1}}\frac{1}{2}\sum_{k,r,s}[(O(V, \alpha)_{krs}-O(V, \alpha_i)_{krs})^2+ (O(V, \alpha)_{krs}-O(V, \alpha_{i+1})_{krs})^2]} \notag 
\end{align}
For any $\alpha_i\leq \alpha\leq \alpha_{i+1}$ on the pixel $(r,s)$, 
\begin{align}
    \Delta_\alpha&\leq \min_{r,s}\min_{P\in \mathbb{P}}\{\sup{\mathbb{U}_{P,r,s}} - \inf{\mathbb{U}_{P,r,s}}\} \notag \\ &\leq \min_{r,s}\min_{P\in \mathbb{P} \mid \floor{\rho(P,\alpha)} = (r,s), \alpha\in\mathbb{U}_{P,r,s}}\sup{\mathbb{U}_{P,r,s}} - \inf{\mathbb{U}_{P,r,s}} \notag \\ &\leq \min_{P\in \mathbb{P} \mid \floor{\rho(P,\alpha)} = (r,s), \alpha\in\mathbb{U}_{P,r,s}}\sup{\mathbb{U}_{P,r,s}} - \inf{\mathbb{U}_{P,r,s}} \notag
\end{align}
Based on Lemma \ref{lemma:find_exact_Delta_re}, we have
$$O(V, \alpha)_{r,s} \in\{O(V, \alpha_i)_{r,s},O(V, \alpha_{i+1})_{r,s}\}$$
i.e., either $O(V, \alpha)_{r,s}=O(V, \alpha_i)_{r,s}$ or $O(V, \alpha)_{r,s}=O(V, \alpha_{i+1})_{r,s}$ holds.
Therefore, for any $k, r, s$, we have
$$(O(V, \alpha)_{krs}-O(V, \alpha_i)_{krs})^2+ (O(V, \alpha)_{krs}-O(V, \alpha_{i+1})_{krs})^2 = (O(V, \alpha_i)_{krs}-O(V, \alpha_{i+1})_{krs})^2$$
Therefore,
\begin{align}
    \centering
    &\max_{\alpha\in\mathcal{S}}\min_{1\leq i \leq N}\left\|\phi(x, \alpha) - \phi(x, \alpha_i)\right\|_2 \notag\\   
    \leq & \max_{1\leq i\leq N-1}\sqrt{\max_{\alpha_i\leq \alpha \leq \alpha_{i+1}}\frac{1}{2}\sum_{k,r,s}[(O(V, \alpha)_{krs}-O(V, \alpha_i)_{krs})^2+ (O(V, \alpha)_{krs}-O(V, \alpha_{i+1})_{krs})^2]} \notag \\
    =&\max_{1\leq i\leq N-1}\sqrt{\frac{1}{2} \sum_{k,r,s}(O(V, \alpha_i)_{krs}-O(V, \alpha_{i+1})_{krs})^2} 
    \notag
\end{align}
So if \begin{equation}
                \begin{gathered} \max_{1\leq i\leq N-1}\sqrt{\frac{1}{2} \sum_{k,r,s}(O(V, \alpha_i)_{krs}-O(V, \alpha_{i+1})_{krs})^2}   < \frac{\sigma}{2}\min_{1\leq i \leq N}\left(\Phi^{-1}\left(p_A^{(i)}\right)-\Phi^{-1}\left(p_B^{(i)}\right)\right).
                \end{gathered}
            \end{equation}
 then it holds that
\begin{equation}
                \label{eq:general_Ms_re1}
                \begin{gathered}
                    M_{\mathcal{S}}:= \max_{\alpha\in\mathcal{S}}\min_{1\leq i \leq N}\left\|\phi(x, \alpha) - \phi(x, \alpha_i)\right\|_2< R:=\frac{\sigma}{2}\min_{1\leq i \leq N}\left(\Phi^{-1}\left(p_A^{(i)}\right)-\Phi^{-1}\left(p_B^{(i)}\right)\right).
                \end{gathered}
            \end{equation}
which concludes the proof based on Theorem \ref{lemma: diff_res}.
\end{proof}

\section{Proofs in Certification via Pixel-wise Smoothing with Lipschitz-approximated Partitions (PwS-L)}
\label{sec:C}
In this section, we show the proof of Lemma \ref{lemma:find_Delta_appro} and Theorem \ref{theorem:diff_res_appro}, where the upper bound of the partitioning interval can be approximated through the Lipschitz property of the projection oracle.
\subsection{Proof on Lemma \ref{lemma:find_Delta_appro}: Approximated upper bound of fully-covered interval}
\begin{lemma}[Approximated upper bound of fully-covered interval, \textbf{restated} from Lemma \ref{lemma:find_Delta_appro}]
\label{lemma:find_Delta_appro_re}
For the  projection from entire 3D point  $V\in\mathcal{V}: \mathbb{P}\times[0, 1]^K$, given the one-axis monotonic position projection function $\rho$ in $l_\infty$ norm over camera motion space,
if the Lipschitz-based interval $\Delta_L^{r,s}$ is defined as
\begin{align}
    \centering
        \label{equ:Delta_L_appro}
   \Delta_L^{r,s}=\min\limits_{P\in \mathbb{P} 
   } \frac{1}{ L_{P}}
   \mid\rho(P,\sup{\mathbb{U}_{P,r,s}})-\rho(P,\inf{\mathbb{U}_{P,r,s}})\mid_\infty
\end{align}
where 
$L_P$ is the  Lipschitz constant for projection function $\rho$ given 3D point $P$,
\begin{align}
    \centering
L_P = \max \{\max_{\alpha\in\mathcal{S}}|\frac{\partial\rho_1(P,\alpha)}{\partial\alpha}|, \max_{\alpha\in\mathcal{S}}|\frac{\partial\rho_2(P,\alpha)}{\partial\alpha}|\}
\end{align}
 then it holds that  $\Delta_L^{r,s} \leq \Delta^{r,s} = \min_{P\in \mathbb{P}}\{\sup{\mathbb{U}_{P,r,s}} - \inf{\mathbb{U}_{P,r,s}}\}  $.
    
\end{lemma}
\begin{proof}
Considering the projection function $\rho$ on $r,s$ with any  3D point $P:\floor{\rho(P,\alpha)}=(r,s)$ at camera motion $\alpha\in\mathcal{S}$ with the least depth, $\mathbb{U}_{P,r,s}\neq \emptyset$
 then the interval $\Delta_L^{r,s}$  satisfies 
 \begin{align*}
     \centering
    \Delta_L^{r,s} &=\min\limits_{P\in \mathbb{P} 
   } \frac{1}{ L_{P}}
   \mid\rho(P,\sup{\mathbb{U}_{P,r,s}})-\rho(P,\inf{\mathbb{U}_{P,r,s}})\mid_\infty\\
    &\leq\min\limits_{\{P\in \mathbb{P} \mid \floor{\rho(P,\alpha)} = (r,s),\alpha\in\mathcal{S} \}} \frac{1}{ L_{P}}
   \mid\rho(P,\sup{\mathbb{U}_{P,r,s}})-\rho(P,\inf{\mathbb{U}_{P,r,s}})\mid_\infty \\
   & \leq \frac{1}{ L_{P}}
   \mid\rho(P,\sup{\mathbb{U}_{P,r,s}})-\rho(P,\inf{\mathbb{U}_{P,r,s}})\mid_\infty
 \end{align*}
For the Lipschitz constant  $L_P$ for projection function $\rho$ with point $P$, we have
\begin{align}
\label{equ:L_P_re}
    \centering
    L_P = &\max \{\max_{\alpha\in\mathcal{S}}|\frac{\partial\rho_1(P,\alpha)}{\partial\alpha}|, \max_{\alpha\in\mathcal{S}}|\frac{\partial\rho_2(P,\alpha)}{\partial\alpha}|\} \notag \\
    =& \max \{\max_{\alpha,\beta\in\mathcal{S}}\frac{|\rho_1(P,\alpha)-\rho_1(P,\beta)|}{|\alpha-\beta|},\max_{\alpha,\beta\in\mathcal{S}}\frac{|\rho_2(P,\alpha)-\rho_2(P,\beta)|}{|\alpha-\beta|}\} \notag \\
    =&\max_{\alpha,\beta\in\mathcal{S}}\max\{\frac{|\rho_1(P,\alpha)-\rho_1(P,\beta)|}{|\alpha-\beta|},\frac{|\rho_2(P,\alpha)-\rho_2(P,\beta)|}{|\alpha-\beta|}\} \notag \\
    =&\max_{\alpha,\beta\in\mathcal{S}}\frac{|\rho(P,\alpha) - \rho(P, \beta|_\infty}{|\alpha-\beta|}
\end{align}
Therefore, we have
\begin{align*}
    \centering
&\max_{\alpha\in\mathcal{S}}\frac{|\rho(P,\alpha+\Delta_L^{r,s}) - \rho(P, \alpha)|_\infty}{\Delta_L^{r,s}} \\ \leq & \max_{\alpha,\beta\in\mathcal{S}}\frac{|\rho(P,\alpha) - \rho(P, \beta)|_\infty}{|\alpha-\beta|} =L_P \\
\leq &  \frac{1}{\Delta_L^{r,s}}
   \mid\rho(P,\sup{\mathbb{U}_{P,r,s}})-\rho(P,\inf{\mathbb{U}_{P,r,s}})\mid_\infty
\end{align*}
 Therefore, for any $u \in \mathcal{S}$
$$\mid\rho(P, u+\Delta_L^{r,s}) - \rho(P, u)\mid_\infty \leq \max_{\alpha\in\mathcal{S}}|\rho(P,\alpha+\Delta_L^{r,s}) - \rho(P, \alpha)|_\infty \leq \mid\rho(P,\sup{\mathbb{U}_{P,r,s}})-\rho(P,\inf{\mathbb{U}_{P,r,s}})\mid_\infty$$
Based on the monotonicity of $\rho$ in $l_\infty$ norm  over $\mathbb{U}_{P,r,s}$ given $P$, let $u=\inf{\mathbb{U}_{P,r,s}}$, we have $\Delta_L^{r,s} \leq \sup{\mathbb{U}_{P,r,s}} - \inf{\mathbb{U}_{P,r,s}}$ for any $P$ and $(r,s)$.
Therefore, for pixel $(r,s)$,
\begin{align}
    \Delta_L^{r,s} \leq \min_{P\in \mathbb{P} \mid \floor{\rho(P,\alpha)} = (r,s), \alpha\in\mathbb{U}_{P,r,s}}\sup{\mathbb{U}_{P,r,s}} - \inf{\mathbb{U}_{P,r,s}}  = \min_{P\in \mathbb{P}}\{\sup{\mathbb{U}_{P,r,s}} - \inf{\mathbb{U}_{P,r,s}}\}  \notag
\end{align}
which concludes the proof.
\end{proof}

\subsection{Proof on Theorem \ref{theorem:diff_res_appro}: Certification with approximated partitions}
\begin{theorem}[Certification with approximated partitions, \textbf{restated} of Theorem \ref{theorem:diff_res_appro}]
    \label{theorem:diff_res_appro_re}
For the projection from entire 3D point cloud $V\in\mathcal{V}: \mathbb{P}\times[0, 1]^K$ through the monotonic position projection function $\rho$ in $l_\infty$ norm over camera motion space, 
let $\phi$ be one-axis rotation or translation with parameters in $\mathcal{S}\subseteq\mathcal{Z}$  and uniformly partitioned $\{\alpha_i\}_{i=1}^N\subseteq\mathcal{S}$ with interval $\Delta_\alpha$, where
\begin{align}
    \centering
   \Delta_\alpha \leq\min \limits_{r,s} \Delta_L^{r,s} &= \min \limits_{r,s} \min\limits_{P\in \mathbb{P} } \frac{1}{ L_{P}}
   \mid\rho(P,\sup{\mathbb{U}_{P,r,s}})  -\rho(P,\inf{\mathbb{U}_{P,r,s}})\mid_\infty \\
 L_P &= \max \{\max_{\alpha\in\mathcal{S}}|\frac{\partial\rho_1(P,\alpha)}{\partial\alpha}|, \max_{\alpha\in\mathcal{S}}|\frac{\partial\rho_2(P,\alpha)}{\partial\alpha}|\}
\end{align}
Let $y_A,y_B\in\mathcal{Y}$, $\varepsilon\sim\mathcal{N}(0, \sigma^2 \mathbb{I}_d)$ and suppose that for any $i$, the $\varepsilon$-smoothed classifier defined by $q(y\mid x;\varepsilon):=\mathbb{E}(p(y\mid x+\varepsilon))$ has class probabilities that satisfy  with top-2 classes $y_A,y_B$ as
$
                 q(y_A\mid \phi(x, \alpha_i); \varepsilon) \geq p_A^{(i)} \geq p_B^{(i)} \geq \max_{y\neq y_A}q(y\mid \phi(x, \alpha_i); \varepsilon)
$,

            then it is guaranteed that $\forall\alpha\in\mathcal{S}\colon y_A = \argmax_y q(y\mid \phi(x, \alpha);\varepsilon)$ if 
            $$ \max_{1\leq i\leq N-1}\sqrt{\frac{1}{2} \sum_{k,r,s}(O(V, \alpha_i)_{krs}-O(V, \alpha_{i+1})_{krs})^2}  < \frac{\sigma}{2}\min_{1\leq i \leq N}\left(\Phi^{-1}\left(p_A^{(i)}\right)-\Phi^{-1}\left(p_B^{(i)}\right)\right).
               $$
\end{theorem}
\begin{proof}
According to Lemma \ref{lemma:find_Delta_appro_re}, at each pixel $(r,s)$
we have 
 \begin{align}
    \Delta_L^{r,s}  \leq \min_{P\in \mathbb{P}}\{\sup{\mathbb{U}_{P,r,s}} - \inf{\mathbb{U}_{P,r,s}}\}   \notag
\end{align}
Therefore, 
$$\Delta_\alpha \leq\min \limits_{r,s} \Delta_L^{r,s} \leq \min_{r,s} \min_{P\in \mathbb{P}}\{\sup{\mathbb{U}_{P,r,s}} - \inf{\mathbb{U}_{P,r,s}}\}  $$
Now applying Theorem \ref{theorem:diff_res_re}, the theorem is directly proved as a corollary.
\end{proof}

\section{Proofs in Certification via Pixel-wise Smoothing with One-Frame Point Cloud as Prior (PwS-OF)}
\label{sec:D}
We finally give the proof of Lemma \ref{lemma:find_Delta_appro_one} and Theorem \ref{theorem:diff_res_appro_one}, where the certification using Lipschitz-based partitioning can be extended to the more general case with only one-frame point cloud known.
\begin{definition}[3D projection from one-frame point cloud with $\delta$-convexity, \textbf{restated} of Definition \ref{def:delta_conv}]
\label{def:delta_conv_re}
Given a one-frame point cloud $\mathbb{P}_1\subset\mathbb{R}^3$, define the  $K$-channel image $x\in\mathcal{X}: [0,1]^K\times\mathbb{Z}^2$  as
\begin{align}
        x_{k,r,s}= O(V, 0)_{k,r,s} =V_{P_1, k},  \text{ where } P_1 \in  \mathbb{P}_1 \text{ s.t. }  \floor{\rho(P_1,\alpha)} = (r,s)
\end{align}
Define  $\delta$-convexity that with the minimal $\delta>0$,  for any new point $P \notin \mathbb{P}_1$ and any $\alpha\in\mathcal{S}$,   there exists $P' \in \mathbb{P}_1$ where
$\mid \rho(P, \alpha) - \rho(P', \alpha)\mid_\infty \leq \delta$, it holds that
$D(P, \alpha) \geq D(P', \alpha)$.
\end{definition}
\subsection{Lemma D.2 and Proof: Lipschitz relaxation for 1-axis translation or rotation}
\begin{lemma}[Lipschitz relaxation for 1-axis translation or rotation]
\label{lemma:find_C_delta}
Given $\delta$-convexity projection from one-frame 3D point cloud $V_1\in\mathcal{V}_1: \mathbb{P}_1\times[0, 1]^K$ and unknown complete point cloud $V\in\mathcal{V}: \mathbb{P}\times[0, 1]^K$,  if the projection function $\rho$ is the 1-axis translation or rotation, for any $P\in\mathbb{P}\backslash\mathbb{P}_1$
there exists a  finite constant $C_\delta\geq0$ such that for any 1-axis camera motion $\alpha\in\mathcal{S}=[-b, b]$, $$L_P\leq \max_{P'\in\mathbb{P}_1}L_{P'} + C_\delta$$
Specifically, for z-axis translation $T_z$ over $\mathcal{S} = [-b ,b]$, $$C_\delta^{T_z} = \max_{(X', Y', Z')\in\mathbb{P}_1}\frac{\delta}{Z'- b}$$
for x-axis translation $T_x$ over $\mathcal{S} = [-b ,b]$, $$C_\delta^{T_x} = 0$$
for y-axis translation $T_y$ over $\mathcal{S} = [-b ,b]$, $$C_\delta^{T_y} = 0$$
for z-axis rotation $R_z$ over $\mathcal{S} = [-b ,b]$, $$C_\delta^{R_z} =  \max\{\frac{f_x}{f_y}, \frac{f_y}{f_x}\}\delta$$
for x-axis rotation $R_x$ over $\mathcal{S} = [-b ,b]$, $$C_\delta^{R_x} =   \frac{\delta^2}{f_y} +\max_{(X', Y', Z')\in\mathbb{P}_1}\max_{\theta=\pm b}\max\{\frac{\delta}{f_y}\frac{f_x|X'| + f_y|Y'\cos\theta + Z'\sin\theta|}{-Y'\sin\theta+Z'\cos\theta}, \frac{2\delta|Y'\cos\theta + Z'\sin\theta|}{-Y'\sin\theta+Z'\cos\theta}\} $$
for y-axis rotation $R_y$ over $\mathcal{S} = [-b ,b]$, $$C_\delta^{R_y} =   \frac{\delta^2}{f_y} +\max_{(X', Y', Z')\in\mathbb{P}_1}\max_{\theta=\pm b}\max\{\frac{2\delta|X'\cos\theta - Z'\sin\theta|}{X'\sin\theta+Z'\cos\theta}, \frac{\delta}{f_x}\frac{f_y|Y'| + f_x|X'\cos\theta - Z'\sin\theta|}{X'\sin\theta+Z'\cos\theta}\} $$
\end{lemma}
\begin{proof}
We prove this lemma through 6 cases of all 1-axis camera motion.
1) For the translation along z-axis,  for any $P=(X, Y,Z)\in\mathbb{P}\backslash\mathbb{P}_1$ and any camera motion $\alpha=\{t_z\}\in[-b,b]$, we have
      \begin{align}
    \centering
    D_{T_z}(P, \alpha)=Z-t_z, \quad \notag
    \rho_{T_z}(P, \alpha)= (\frac{f_xX+c_x(Z-t_z)}{Z-t_z},
 \frac{f_yY+c_y(Z-t_z)}{Z-t_z})
\end{align}
\begin{align}
    \centering
L_P & = \max \{\max_{\alpha\in\mathcal{S}}|\frac{\partial\rho_1(P,\alpha)}{\partial\alpha}|, \max_{\alpha\in\mathcal{S}}|\frac{\partial\rho_2(P,\alpha)}{\partial\alpha}|\} \notag \\
& = \max\{\max_{t_z\in[-b,b]}\frac{f_x|X|}{(Z-t_z)^2}, \max_{t_z\in[-b,b]}
 \frac{f_y|Y|}{(Z-t_z)^2}\}\notag\\
 & = \max\{\frac{f_x|X|}{(Z-b)^2}, 
 \frac{f_y|Y|}{(Z-b)^2}\} \quad (\text{by} ~ a<b<Z)\notag
\end{align}
According to the definition of  $\delta$-convexity, there exists $P' = (X', Y', Z') \in \mathbb{P}_1$ where
$\mid \rho(P, \alpha) - \rho(P', \alpha)\mid_\infty \leq \delta$, it holds that
$$D(P, \alpha) \geq D(P', \alpha)$$
So we have for any $t_z\in[-b,b]$
$$|\frac{f_xX}{Z-t_z} - \frac{f_xX'}{Z'-t_z}|\leq\delta, |\frac{f_yY}{Z-t_z} - \frac{f_yY'}{Z'-t_z}|\leq\delta, Z'-t_z<Z-t_z$$
\begin{align}
    \centering
L_P& = \max\{\frac{f_x|X|}{(Z-b)^2}, 
 \frac{f_y|Y|}{(Z-b)^2}\} \notag \\
 & \leq \max\{\frac{1}{Z-b}(\frac{f_x|X'|}{Z'-b}+\delta), 
 \frac{1}{Z-b}(\frac{f_y|Y'|}{Z'-b}+\delta)\}\notag \\
  & \leq \max\{\frac{1}{Z'-b}(\frac{f_x|X'|}{Z'-b}+\delta), 
 \frac{1}{Z'-b}(\frac{f_y|Y'|}{Z'-b}+\delta)\} \notag \\
   & =\max\{\frac{f_x|X'|}{(Z'-b)^2}, 
 \frac{f_y|Y'|}{(Z'-b)^2}\} + \frac{\delta}{Z'-b} \notag \\
 & \leq L_{P'} + \max_{(X', Y', Z')\in\mathbb{P}_1}\frac{\delta}{Z'-b} \notag \\
 & \leq \max_{P'\in\mathbb{P}_1}L_{P'} + C_\delta^{T_z} \notag
\end{align}
where $$C_\delta^{T_z} = \max_{(X', Y', Z')\in\mathbb{P}_1}\frac{\delta}{Z'- b}$$

2) For the translation along x-axis,  for any $P=(X, Y,Z)\in\mathbb{P}\backslash\mathbb{P}_1$ and any camera motion $\alpha=\{t_x\}\in[-b,b]$, we have
      \begin{align}
    \centering
    D_{T_x}(P, \alpha)=Z, \quad \notag
    \rho_{T_x}(P, \alpha)=(\frac{f_x(X-t_x)+c_xZ}{Z},
 \frac{f_yY+c_yZ}{Z})
\end{align}
\begin{align}
    \centering
L_P & = \max \{\max_{\alpha\in\mathcal{S}}|\frac{\partial\rho_1(P,\alpha)}{\partial\alpha}|, \max_{\alpha\in\mathcal{S}}|\frac{\partial\rho_2(P,\alpha)}{\partial\alpha}|\} \notag \\
& = \max\{\max_{t_x\in[-b,b]} \frac{f_x}{Z} , 0\}\notag\\
 & =  \frac{f_x}{Z}\notag
\end{align}
According to the definition of  $\delta$-convexity, there exists $P' = (X', Y', Z') \in \mathbb{P}_1$ where
$\mid \rho(P, \alpha) - \rho(P', \alpha)\mid_\infty \leq \delta$, it holds that
$$D(P, \alpha) \geq D(P', \alpha)$$
So we have  $Z' < Z$
\begin{align}
    \centering
L_P& = \frac{f_x}{Z} \leq \frac{f_x}{Z'}  = L_{P'} + 0\notag \\
 & \leq \max_{P'\in\mathbb{P}_1}L_{P'} + C_\delta^{T_x} \notag
\end{align}
where $$ C_\delta^{T_x} = 0$$

3) For the translation along y-axis,  for any $P=(X, Y,Z)\in\mathbb{P}\backslash\mathbb{P}_1$ and any camera motion $\alpha=\{t_y\}\in[-b,b]$, we have
      \begin{align}
    \centering
   D_{T_y}(P, \alpha)=Z, \quad \notag
    \rho_{T_y}(P, \alpha)= (\frac{f_xX+c_xZ}{Z},
\frac{f_y(Y-t_y)+c_yZ}{Z})
\end{align}
\begin{align}
    \centering
L_P & = \max \{\max_{\alpha\in\mathcal{S}}|\frac{\partial\rho_1(P,\alpha)}{\partial\alpha}|, \max_{\alpha\in\mathcal{S}}|\frac{\partial\rho_2(P,\alpha)}{\partial\alpha}|\} \notag \\
& = \max\{0, \max_{t_y\in[-b,b]} \frac{f_y}{Z}\}\notag\\
 & =  \frac{f_y}{Z}\notag
\end{align}
According to the definition of  $\delta$-convexity, there exists $P' = (X', Y', Z') \in \mathbb{P}_1$ where
$\mid \rho(P, \alpha) - \rho(P', \alpha)\mid_\infty \leq \delta$, it holds that
$$D(P, \alpha) \geq D(P', \alpha)$$
So we have  $Z'< Z$
\begin{align}
    \centering
L_P& = \frac{f_y}{Z} \leq \frac{f_y}{Z'}  = L_{P'} + 0\notag \\
 & \leq \max_{P'\in\mathbb{P}_1}L_{P'} + C_\delta^{T_y} \notag
\end{align}
where $$ C_\delta^{T_y} = 0$$
        
4) For the rotation along z-axis,  for any $P=(X, Y,Z)\in\mathbb{P}\backslash\mathbb{P}_1$ and any camera motion $\alpha=\{\theta\}\in[-b,b]$, we have
      \begin{align}
    \centering
    D_{R_z}(P, \alpha)=Z, \quad \notag
    \rho_{R_z}(P, \alpha)= (\frac{f_x\cos\theta X+f_x \sin\theta  Y}{Z}+c_x,
\frac{f_y\cos\theta Y+f_y \sin\theta  X}{Z}+c_y)
\end{align}
\begin{align}
    \centering
L_P & = \max \{\max_{\alpha\in\mathcal{S}}|\frac{\partial\rho_1(P,\alpha)}{\partial\alpha}|, \max_{\alpha\in\mathcal{S}}|\frac{\partial\rho_2(P,\alpha)}{\partial\alpha}|\} \notag \\
& = \max\{\max_{\theta\in[-b,b]}\frac{|-X\sin\theta + Y\cos\theta|}{Z} f_x, \max_{\theta\in[-b,b]}
 \frac{|-Y\sin\theta + X\cos\theta|}{Z} f_y\}\notag\\
& = \max\{\frac{|-X\sin\theta_1 + Y\cos\theta_1|}{Z} f_x,
 \frac{|-Y\sin\theta_2 + X\cos\theta_2|}{Z} f_y\}\notag
\end{align}
where $$\theta_1 = \argmax_{\theta\in[-b,b]} \frac{|-X\sin\theta + Y\cos\theta|}{Z} f_x, \theta_2 = \argmax_{\theta\in[-b,b]}
 \frac{|-Y\sin\theta + X\cos\theta|}{Z} f_y$$
According to the definition of  $\delta$-convexity, there exists $P' = (X', Y', Z') \in \mathbb{P}_1$ where
$\mid \rho(P, \alpha) - \rho(P', \alpha)\mid_\infty \leq \delta$, it holds that
$$D(P, \alpha) \geq D(P', \alpha)$$
So we have $Z' < Z$ and  for any $\theta\in[-b,b]$
$$|\frac{f_x\cos\theta X+f_x \sin\theta  Y}{Z} - \frac{f_x\cos\theta X'+f_x \sin\theta  Y'}{Z'}|\leq\delta, |\frac{f_y\cos\theta Y+f_y \sin\theta  X}{Z} -\frac{f_y\cos\theta Y'+f_y \sin\theta  X'}{Z'}|\leq\delta$$
\begin{align}
    \centering
L_P
 =& \max\{\frac{|-X\sin\theta_1 + Y\cos\theta_1|}{Z} f_x,
 \frac{|-Y\sin\theta_2 + X\cos\theta_2|}{Z} f_y\}\notag \\
  \leq &  \max\{\frac{|-X'\sin\theta_1 + Y'\cos\theta_1|}{Z'} f_x,
 \frac{|-Y'\sin\theta_2 + X'\cos\theta_2|}{Z'} f_y\} \notag\\&+ \max\{|-\sin\theta_1(\frac{X}{Z} - \frac{X'}{Z'}) + \cos\theta_1(\frac{Y}{Z} - \frac{Y'}{Z'})| f_x,
|-\sin\theta_2(\frac{Y}{Z} - \frac{Y'}{Z'}) + \cos\theta_2(\frac{X}{Z} - \frac{X'}{Z'})| f_y\} \notag \\
 \leq&  \max\{\max_{\theta\in[-b,b]}\frac{|-X'\sin\theta + Y'\cos\theta|}{Z'} f_x, \max_{\theta\in[-b,b]}
 \frac{|-Y'\sin\theta + X'\cos\theta|}{Z'} f_y\} \notag\\&+  \max\{|\frac{f_y\cos(-\theta_1) Y+f_y \sin(-\theta_1)  X}{Z} -\frac{f_y\cos(-\theta_1) Y'+f_y \sin(-\theta_1)  X'}{Z'}| \frac{f_x}{f_y},\notag \\
&|\frac{f_x\cos(-\theta_2) X+f_x \sin(-\theta_2)  Y}{Z} - \frac{f_x\cos(-\theta_2) X'+f_x \sin(-\theta_2)  Y'}{Z'}| \frac{f_y}{f_x}\} \notag \\
  \leq& L_{P'} + \max\{\frac{f_x}{f_y}, \frac{f_y}{f_x}\}\delta \notag \\
  \leq& \max_{P'\in\mathbb{P}_1}L_{P'} + C_\delta^{R_z} \notag 
\end{align}
where $$C_\delta^{R_z} =  \max\{\frac{f_x}{f_y}, \frac{f_y}{f_x}\}\delta$$
        
5) For the rotation along x-axis,  for any $P=(X, Y,Z)\in\mathbb{P}\backslash\mathbb{P}_1$ and any camera motion $\alpha=\{\theta\}\in[-b,b]$, we have
      \begin{align}
    \centering
     D_{R_x}(P, \alpha)&=-\sin\theta Y+\cos\theta Z \notag\\
    \rho_{R_x}(P, \alpha)&=(\frac{f_x X}{-Y\sin\theta +Z\cos\theta }+c_x,
     \notag
 \frac{Y\cos\theta +Z \sin\theta}{-Y\sin\theta +Z\cos\theta }f_y +c_y) 
\end{align}
\begin{align}
    \centering
L_P & = \max \{\max_{\alpha\in\mathcal{S}}|\frac{\partial\rho_1(P,\alpha)}{\partial\alpha}|, \max_{\alpha\in\mathcal{S}}|\frac{\partial\rho_2(P,\alpha)}{\partial\alpha}|\} \notag \\
& = \max\{\max_{\theta\in[-b,b]}\frac{|(Y\cos\theta+Z\sin\theta)X|}{(-Y\sin\theta+Z\cos\theta)^2} f_x, \max_{\theta\in[-b,b]}
\frac{Y^2+Z^2}{(-Y\sin\theta+Z\cos\theta)^2} f_y\}\notag\\
& = \max\{\frac{|(Y\cos\theta_1+Z\sin\theta_1)X|}{(-Y\sin\theta_1+Z\cos\theta_1)^2} f_x,
 \frac{Y^2+Z^2}{(-Y\sin\theta_2+Z\cos\theta_2)^2} f_y\}\notag
\end{align}
where $$\theta_1 = \argmax_{\theta\in[-b,b]} \frac{|(Y\cos\theta+Z\sin\theta)X|}{(-Y\sin\theta+Z\cos\theta)^2} f_x, \theta_2 = \argmax_{\theta\in[-b,b]}
 \frac{Y^2+Z^2}{(-Y\sin\theta+Z\cos\theta)^2} f_y$$
According to the definition of  $\delta$-convexity, there exists $P' = (X', Y', Z') \in \mathbb{P}_1$ where
$\mid \rho(P, \alpha) - \rho(P', \alpha)\mid_\infty \leq \delta$, it holds that
$$D(P, \alpha) \geq D(P', \alpha)$$
So we have,  for any $\theta\in[-b,b]$, 
$$0<-\sin\theta Y'+\cos\theta Z' < -\sin\theta Y+\cos\theta Z$$ 
$$|\frac{f_x X}{-Y\sin\theta +Z\cos\theta} - \frac{f_x X'}{-Y'\sin\theta +Z'\cos\theta}|\leq\delta, |\frac{Y\cos\theta +Z \sin\theta}{-Y\sin\theta +Z\cos\theta }f_y -\frac{Y'\cos\theta +Z' \sin\theta}{-Y'\sin\theta +Z'\cos\theta }f_y|\leq\delta$$
\begin{align}
    \centering
L_P
 =& \max\{\frac{|(Y\cos\theta_1+Z\sin\theta_1)X|}{(-Y\sin\theta_1+Z\cos\theta_1)^2} f_x,
 \frac{Y^2+Z^2}{(-Y\sin\theta_2+Z\cos\theta_2)^2} f_y\}\notag \\
 \leq & \max\{\frac{|(Y'\cos\theta_1+Z'\sin\theta_1)X'|}{(-Y'\sin\theta_1+Z'\cos\theta_1)^2} f_x,
 \frac{Y'^2+Z'^2}{(-Y'\sin\theta_2+Z'\cos\theta_2)^2} f_y\} \notag \\ 
 &+  \max\{\frac{\delta^2}{f_y} + \frac{\delta}{f_y}\frac{f_x|X'| + f_y|Y'\cos\theta_1 + Z'\sin\theta_1|}{-Y'\sin\theta_1+Z'\cos\theta_1}, \frac{\delta^2}{f_y}+\frac{2\delta|Y'\cos\theta_2 + Z'\sin\theta_2|}{-Y'\sin\theta_2+Z'\cos\theta_2}\}
 \notag \\
 \leq &  \max\{\max_{\theta\in[-b,b]}\frac{|(Y'\cos\theta+Z'\sin\theta)X|}{(-Y'\sin\theta+Z'\cos\theta)^2} f_x, \max_{\theta\in[-b,b]}
\frac{Y'^2+Z'^2}{(-Y'\sin\theta+Z'\cos\theta)^2} f_y\} \notag \\ &+  \max\{\max_{\theta\in[-b,b]} \frac{\delta^2}{f_y} + \frac{\delta}{f_y}\frac{f_x|X'| + f_y|Y'\cos\theta + Z'\sin\theta|}{-Y'\sin\theta+Z'\cos\theta}, \max_{\theta\in[-b,b]}  \frac{\delta^2}{f_y}+\frac{2\delta|Y'\cos\theta + Z'\sin\theta|}{-Y'\sin\theta+Z'\cos\theta}\}\notag
\end{align}
It is easy to find that, 
$$\frac{\partial \frac{ |X'|}{-Y'\sin\theta +Z'\cos\theta} }{\partial\theta} = \frac{ |X|(Y'\cos\theta+Z'\sin\theta)}{(-Y'\sin\theta +Z'\cos\theta)^2}, \frac{\partial \frac{ Y'\cos\theta + Z'\sin\theta}{-Y'\sin\theta +Z'\cos\theta} }{\partial\theta} = \frac{ Y'^2+Z'^2}{(-Y'\sin\theta +Z'\cos\theta)^2}$$
So the functions below are increasing if $ Y'\cos\theta+Z'\sin\theta > 0$, and vice versa. $$\max_{\theta\in[-b,b]} \frac{|X'|}{-Y'\sin\theta+Z'\cos\theta} = \max\{\frac{|X'|}{-Y'\sin b+Z'\cos b}, \frac{|X'|}{Y'\sin b+Z'\cos b}\}$$
$$\max_{\theta\in[-b,b]} \frac{|Y'\cos\theta + Z'\sin\theta|}{-Y'\sin\theta+Z'\cos\theta} = \max\{\frac{Y'\cos b + Z'\sin b}{-Y'\sin b+Z'\cos b}, \frac{Y'\cos b - Z'\sin b}{Y'\sin b+Z'\cos b}\}$$
\begin{align}
    \centering
    L_P \leq &   \max\{\max_{\theta\in[-b,b]}\frac{|(Y'\cos\theta+Z'\sin\theta)X|}{(-Y'\sin\theta+Z'\cos\theta)^2} f_x, \max_{\theta\in[-b,b]}
\frac{Y'^2+Z'^2}{(-Y'\sin\theta+Z'\cos\theta)^2} f_y\} \notag \\ &+  \max\{\max_{\theta\in[-b,b]} \frac{\delta^2}{f_y} + \frac{\delta}{f_y}\frac{f_x|X'| + f_y|Y'\cos\theta + Z'\sin\theta|}{-Y'\sin\theta+Z'\cos\theta}, \max_{\theta\in[-b,b]}  \frac{\delta^2}{f_y}+\frac{2\delta|Y'\cos\theta + Z'\sin\theta|}{-Y'\sin\theta+Z'\cos\theta}\}\notag \\
 \leq&  L_{P'}  + \frac{\delta^2}{f_y} +\max_{(X', Y', Z')\in\mathbb{P}_1}\max_{\theta=\pm b}\max\{\frac{\delta}{f_y}\frac{f_x|X'| + f_y|Y'\cos\theta + Z'\sin\theta|}{-Y'\sin\theta+Z'\cos\theta}, \frac{2\delta|Y'\cos\theta + Z'\sin\theta|}{-Y'\sin\theta+Z'\cos\theta}\} \notag \\
  \leq& \max_{P'\in\mathbb{P}_1}L_{P'} + C_\delta^{R_x} \notag 
\end{align}
where $$C_\delta^{R_x} =   \frac{\delta^2}{f_y} +\max_{(X', Y', Z')\in\mathbb{P}_1}\max_{\theta=\pm b}\max\{\frac{\delta}{f_y}\frac{f_x|X'| + f_y|Y'\cos\theta + Z'\sin\theta|}{-Y'\sin\theta+Z'\cos\theta}, \frac{2\delta|Y'\cos\theta + Z'\sin\theta|}{-Y'\sin\theta+Z'\cos\theta}\} $$

6) For the rotation along y-axis,  for any $P=(X, Y,Z)\in\mathbb{P}\backslash\mathbb{P}_1$ and any camera motion $\alpha=\{\theta\}\in[-b,b]$, we have
      \begin{align}
    \centering
    D_{R_y}(P, \alpha)&=\sin\theta X+\cos\theta Z \notag
\\
    \rho_{R_y}(P, \alpha)&=(\frac{X\cos\theta- Z\sin\theta}{X\sin\theta + Z\cos\theta}f_x + c_x,\notag
 \frac{f_y Y}{X\sin\theta + Z\cos\theta} +c_y)
\end{align}
\begin{align}
    \centering
L_P & = \max \{\max_{\alpha\in\mathcal{S}}|\frac{\partial\rho_1(P,\alpha)}{\partial\alpha}|, \max_{\alpha\in\mathcal{S}}|\frac{\partial\rho_2(P,\alpha)}{\partial\alpha}|\} \notag \\
& = \max\{\max_{\theta\in[-b,b]} \frac{X^2+Z^2}{(X\sin\theta+Z\cos\theta)^2} f_x, \max_{\theta\in[-b,b]}
\frac{|(X\cos\theta-Z\sin\theta)Y|}{(X\sin\theta+Z\cos\theta)^2} f_y\}\notag\\
& = \max\{ \frac{X^2+Z^2}{(X\sin\theta_1+Z\cos\theta_1)^2} f_x,
\frac{|(X\cos\theta_2-Z\sin\theta_2)Y|}{(X\sin\theta_2+Z\cos\theta_2)^2} f_y\}\notag
\end{align}
where $$\theta_1 = \argmax_{\theta\in[-b,b]} \max_{\theta\in[-b,b]} \frac{X^2+Z^2}{(X\sin\theta+Z\cos\theta)^2} f_x, \theta_2 = \argmax_{\theta\in[-b,b]}
 \frac{|(X\cos\theta-Z\sin\theta)Y|}{(X\sin\theta+Z\cos\theta)^2} f_y$$
According to the definition of  $\delta$-convexity, there exists $P' = (X', Y', Z') \in \mathbb{P}_1$ where
$\mid \rho(P, \alpha) - \rho(P', \alpha)\mid_\infty \leq \delta$, it holds that
$$D(P, \alpha) \geq D(P', \alpha)$$
So we have,  for any $\theta\in[-b,b]$, 
$$0<\sin\theta X'+\cos\theta Z' < \sin\theta X+\cos\theta Z$$ 
$$|\frac{X\cos\theta- Z\sin\theta}{X\sin\theta + Z\cos\theta}f_x - \frac{X'\cos\theta- Z'\sin\theta}{X'\sin\theta + Z'\cos\theta}f_x|\leq\delta, |\frac{f_y Y}{X\sin\theta + Z\cos\theta} -\frac{f_y Y'}{X'\sin\theta + Z'\cos\theta}|\leq\delta$$
\begin{align}
    \centering
L_P
 =& \max\{ \frac{X^2+Z^2}{(X\sin\theta_1+Z\cos\theta_1)^2} f_x,
\frac{|(X\cos\theta_2-Z\sin\theta_2)Y|}{(X\sin\theta_2+Z\cos\theta_2)^2} f_y\}\notag \\
 \leq & \max\{ \frac{X'^2+Z'^2}{(X'\sin\theta_1+Z'\cos\theta_1)^2} f_x,
\frac{|(X'\cos\theta_2-Z'\sin\theta_2)Y'|}{(X'\sin\theta_2+Z'\cos\theta_2)^2} f_y\} \notag \\ 
 &+  \max\{\frac{\delta^2}{f_x}+\frac{2\delta|X'\cos\theta_1 - Z'\sin\theta_1|}{X'\sin\theta_1+Z'\cos\theta_1}, \frac{\delta^2}{f_x} + \frac{\delta}{f_x}\frac{f_y|Y'| + f_x|X'\cos\theta_2 - Z'\sin\theta_2|}{X'\sin\theta_2+Z'\cos\theta_2}\}
 \notag \\
 \leq &  \max\{\max_{\theta\in[-b,b]}\frac{X^2+Z^2}{(X\sin\theta+Z\cos\theta)^2} f_x, \max_{\theta\in[-b,b]}
\frac{|(X\cos\theta-Z\sin\theta)Y|}{(X\sin\theta+Z\cos\theta)^2} f_y\} \notag \\ &+  \max\{\max_{\theta\in[-b,b]} \frac{\delta^2}{f_x}+\frac{2\delta|X'\cos\theta - Z'\sin\theta|}{X'\sin\theta+Z'\cos\theta}, \max_{\theta\in[-b,b]}  \frac{\delta^2}{f_x} + \frac{\delta}{f_x}\frac{f_y|Y'| + f_x|X'\cos\theta - Z'\sin\theta|}{X'\sin\theta+Z'\cos\theta}\}\notag
\end{align}
It is easy to find that, 
$$\frac{\partial \frac{ |Y'|}{X'\sin\theta +Z'\cos\theta} }{\partial\theta} = \frac{ |X|(X'\cos\theta-Z'\sin\theta)}{(X'\sin\theta +Z'\cos\theta)^2}, \frac{\partial \frac{ X'\cos\theta - Z'\sin\theta}{X'\sin\theta +Z'\cos\theta} }{\partial\theta} = \frac{ X'^2+Z'^2}{(X'\sin\theta +Z'\cos\theta)^2}$$
So the functions below are increasing if $ X'\cos\theta-Z'\sin\theta > 0$, and vice versa. $$\max_{\theta\in[-b,b]} \frac{ |Y'|}{X'\sin\theta +Z'\cos\theta} = \max\{\frac{ |Y'|}{X'\sin b +Z'\cos b}, \frac{ |Y'|}{-X'\sin b  +Z'\cos b}\}$$
$$\max_{\theta\in[-b,b]} \frac{ X'\cos\theta - Z'\sin\theta}{X'\sin\theta +Z'\cos\theta} = \max\{\frac{ X'\cos b - Z'\sin b}{X'\sin b +Z'\cos b}, \frac{ X'\cos b + Z'\sin b}{-X'\sin b +Z'\cos b}\}$$
\begin{align}
    \centering
    L_P  \leq &  \max\{\max_{\theta\in[-b,b]}\frac{X^2+Z^2}{(X\sin\theta+Z\cos\theta)^2} f_x, \max_{\theta\in[-b,b]}
\frac{|(X\cos\theta-Z\sin\theta)Y|}{(X\sin\theta+Z\cos\theta)^2} f_y\} \notag \\ &+  \max\{\max_{\theta\in[-b,b]} \frac{\delta^2}{f_x}+\frac{2\delta|X'\cos\theta - Z'\sin\theta|}{X'\sin\theta+Z'\cos\theta}, \max_{\theta\in[-b,b]}  \frac{\delta^2}{f_x} + \frac{\delta}{f_x}\frac{f_y|Y'| + f_x|X'\cos\theta - Z'\sin\theta|}{X'\sin\theta+Z'\cos\theta}\}\notag \\
 \leq&  L_{P'}  + \frac{\delta^2}{f_x} +\max_{(X', Y', Z')\in\mathbb{P}_1}\max_{\theta=\pm b}\max\{\frac{2\delta|X'\cos\theta - Z'\sin\theta|}{X'\sin\theta+Z'\cos\theta}, \frac{\delta}{f_x}\frac{f_y|Y'| + f_x|X'\cos\theta - Z'\sin\theta|}{X'\sin\theta+Z'\cos\theta}\} \notag \\
  \leq& \max_{P'\in\mathbb{P}_1}L_{P'} + C_\delta^{R_y} \notag 
\end{align}
where $$C_\delta^{R_y} =   \frac{\delta^2}{f_y} +\max_{(X', Y', Z')\in\mathbb{P}_1}\max_{\theta=\pm b}\max\{\frac{2\delta|X'\cos\theta - Z'\sin\theta|}{X'\sin\theta+Z'\cos\theta}, \frac{\delta}{f_x}\frac{f_y|Y'| + f_x|X'\cos\theta - Z'\sin\theta|}{X'\sin\theta+Z'\cos\theta}\} $$
Therefore, there exists a  finite constant $C_\delta\geq0$ such that for any 1-axis camera motion $\alpha\in\mathcal{S}=[-b, b]$, $L_P\leq \max_{P'\in\mathbb{P}_1}L_{P'} + C_\delta$, and finding all such finite constants concludes the proof.
\end{proof}

\subsection{Proof on Lemma \ref{lemma:find_Delta_appro_one}: Approximated upper bound of the interval from one-frame point cloud}
\begin{lemma}[Approximated upper bound of interval from one-frame point cloud, \textbf{restated} of Lemma \ref{lemma:find_Delta_appro_one}]
\label{lemma:find_Delta_appro_one_re}
Given the projection from one-frame 3D point cloud $V_1\in\mathcal{V}_1: \mathbb{P}_1\times[0, 1]^K$ and unknown entire point cloud $V\in\mathcal{V}: \mathbb{P}\times[0, 1]^K$ with $\delta$-convexity, if the one-axis position projection function $\rho$ is monotonic in $l_\infty$ norm over camera motion space, there exists a  finite constant $C_\delta\geq0$ such that 
if  the  interval $\Delta_{OF}^{r,s}$ is defined as,
\begin{align}
    \Delta_{OF}^{r,s} = \min \limits_{P_1\in\mathbb{P}_1}\frac{1}{ L_{P_1} + C_\delta}\min_{P_1\in\mathbb{P}_1}( \mid\rho(P_1,\sup{\mathbb{U}_{P_1,r,s}})-\rho(P_1,\inf{\mathbb{U}_{P_1,r,s}})\mid_\infty-2\delta)
\end{align}
where 
$L_{P_1}$ is the  Lipschitz constant for projection function $\rho$ given 3D point $P_1$,
\begin{align}
    \centering
L_{P_1} = \max \{\max_{\alpha\in\mathcal{S}}|\frac{\partial\rho_1(P_1,\alpha)}{\partial\alpha}|, \max_{\alpha\in\mathcal{S}}|\frac{\partial\rho_2(P_1,\alpha)}{\partial\alpha}|\}
\end{align}
 then for any $P\in\mathbb{P}$, it holds that $ \Delta_{OF}^{r,s}  \leq \Delta^{r,s} = \min_{P\in \mathbb{P}}\{\sup{\mathbb{U}_{P,r,s}} - \inf{\mathbb{U}_{P,r,s}}\}$
\end{lemma}
\begin{proof}
Since we do not know the entire point cloud as an oracle, we categorize $P\in \mathbb{P}$ within known 3D point cloud $\mathbb{P}_1$ and unknown 3D points $\mathbb{P}\backslash\mathbb{P}_1$ and discuss the two cases respectively.

(1) If   $P \in \mathbb{P}_1$, because $\mathbb{P}_1$ is reconstructed from one image, for any $\alpha\in\mathcal{S}, (r,s) \in\mathbb{G}$ the projection equation $\floor{\rho(P,\alpha)}=(r,s)$ has only one unique solution $P$, i.e.
$$\{P\in \mathbb{P}_1 \mid \floor{\rho(P,\alpha)} = (r,s),\alpha\in\mathcal{S} \} = \{P_1\in \mathbb{P}_1 \}$$
Besides, with $C_\delta \geq 0, \delta > 0$,  we have 
\begin{align}
    \centering
    \Delta_{OF}^{r,s} &= \min \limits_{P_1\in\mathbb{P}_1}\frac{1}{ L_{P_1} + C_\delta}\min_{P_1\in\mathbb{P}_1}( \mid\rho(P_1,\sup{\mathbb{U}_{P_1,r,s}})-\rho(P_1,\inf{\mathbb{U}_{P_1,r,s}})\mid_\infty-2\delta) \notag \\
    &\leq \min \limits_{P_1\in\mathbb{P}_1}\frac{1}{ L_{P_1} + C_\delta}( \mid\rho(P_1,\sup{\mathbb{U}_{P_1,r,s}})-\rho(P_1,\inf{\mathbb{U}_{P_1,r,s}})\mid_\infty-2\delta) \\ \notag
    & < \min\limits_{P\in \mathbb{P}_1} \frac{1}{ L_{P}} \mid\rho(P,\sup{\mathbb{U}_{P,r,s}})-\rho(P,\inf{\mathbb{U}_{P,r,s}})\mid_\infty \notag
\end{align}
Then directly based on Lemma \ref{lemma:find_Delta_appro_re}, we have
 for any $P\in\mathbb{P}_1$, at pixel $(r,s)$,
$$\Delta_{OF}^{r,s} \leq   \min_{P\in \mathbb{P}}\{\sup{\mathbb{U}_{P,r,s}} - \inf{\mathbb{U}_{P,r,s}}\}  $$
the proof is concluded.

(2) If $P\in\mathbb{P}\backslash\mathbb{P}_1$, suppose it is projected to  on $r,s$ through projection function $\rho$ from 3D point $P:\floor{\rho(P,\alpha)}=(r,s)$ over camera motion set $\mathbb{U}_{P,r,s}\neq \emptyset$,
based on Lemma \ref{lemma:find_C_delta}, for the 1-axis translation or rotation projective function $\rho$, there exists a constant $C_\delta\geq0$ such that 
$$L_P\leq \max_{P_1\in\mathbb{P}_1} L_{P_1} + C_\delta$$
 Therefore, the interval $\Delta_{OF}^{r,s}$  satisfies 
 \begin{align*}
     \centering
    \Delta_{OF}^{r,s} &= \min \limits_{P_1\in\mathbb{P}_1}\frac{1}{ L_{P_1} + C_\delta}\min_{P_1\in\mathbb{P}_1}( \mid\rho(P_1,\sup{\mathbb{U}_{P_1,r,s}})-\rho(P_1,\inf{\mathbb{U}_{P_1,r,s}})\mid_\infty-2\delta) \\
    &\leq \frac{1}{ L_{P}}\min_{P_1\in\mathbb{P}_1}( \mid\rho(P_1,\sup{\mathbb{U}_{P_1,r,s}})-\rho(P_1,\inf{\mathbb{U}_{P_1,r,s}})\mid_\infty-2\delta) 
 \end{align*}
For the Lipschitz constant  $L_P$ for projection function $\rho$ with point $P\in\mathbb{P}\backslash\mathbb{P}_1$, we have
\begin{align}
    \centering
    L_P = &\max \{\max_{\alpha\in\mathcal{S}}|\frac{\partial\rho_1(P,\alpha)}{\partial\alpha}|, \max_{\alpha\in\mathcal{S}}|\frac{\partial\rho_2(P,\alpha)}{\partial\alpha}|\} \notag \\
    =& \max \{\max_{\alpha,\beta\in\mathcal{S}}\frac{|\rho_1(P,\alpha)-\rho_1(P,\beta)|}{|\alpha-\beta|},\max_{\alpha,\beta\in\mathcal{S}}\frac{|\rho_2(P,\alpha)-\rho_2(P,\beta)|}{|\alpha-\beta|}\} \notag \\
    =&\max_{\alpha,\beta\in\mathcal{S}}\max\{\frac{|\rho_1(P,\alpha)-\rho_1(P,\beta)|}{|\alpha-\beta|},\frac{|\rho_2(P,\alpha)-\rho_2(P,\beta)|}{|\alpha-\beta|}\} \notag \\
    =&\max_{\alpha,\beta\in\mathcal{S}}\frac{|\rho(P,\alpha) - \rho(P, \beta|_\infty}{|\alpha-\beta|}
\end{align}
Therefore, we have
\begin{align*}
    \centering
&\max_{\alpha\in\mathcal{S}}\frac{|\rho(P,\alpha+\Delta_{OF}^{r,s}) - \rho(P, \alpha)|_\infty}{\Delta_{OF}^{r,s}} \\ \leq & \max_{\alpha,\beta\in\mathcal{S}}\frac{|\rho(P,\alpha) - \rho(P, \beta)|_\infty}{|\alpha-\beta|} =L_P \\
\leq &  \frac{1}{\Delta_{OF}^{r,s}}
\min_{P_1\in\mathbb{P}_1}( \mid\rho(P_1,\sup{\mathbb{U}_{P_1,r,s}})-\rho(P_1,\inf{\mathbb{U}_{P_1,r,s}})\mid_\infty-2\delta)
\end{align*}
From the Definition \ref{def:delta_conv_re}, when the camera motion is at $\sup{\mathbb{U}_{P,r,s}}$, there exists $\mathbb{P}_{sup} \subset \mathbb{P}_1$ such that 
$$\forall P'_{sup}\in \mathbb{P}_{sup}, \mid \rho(P, \sup{\mathbb{U}_{P,r,s}}) - \rho(P'_{sup}, \sup{\mathbb{U}_{P,r,s}})\mid_\infty \leq \delta$$
Similarly when the camera motion is at $\inf{\mathbb{U}_{P,r,s}}$,  there exists $\mathbb{P}_{inf} \subset \mathbb{P}_1$ such that
$$\forall P'_{inf}\in \mathbb{P}_{inf}, \mid \rho(P, \inf{\mathbb{U}_{P,r,s}}) - \rho(P'_{inf}, \inf{\mathbb{U}_{P,r,s}})\mid_\infty \leq \delta$$
Now choose $P'\in \mathbb{P}_{inf}\cap\mathbb{P}_{sup}$ such that $$\mid\rho(P',\sup{\mathbb{U}_{P,r,s}})-\rho(P',\inf{\mathbb{U}_{P,r,s}})\mid_\infty \geq 2\delta $$
Therefore, based on the Absolute Value Inequalities, we have
\begin{align*}
    \centering
&\mid\rho(P,\sup{\mathbb{U}_{P,r,s}})-\rho(P,\inf{\mathbb{U}_{P,r,s}})\mid_\infty \\  \geq & \mid\mid\rho(P',\sup{\mathbb{U}_{P,r,s}})-\rho(P',\inf{\mathbb{U}_{P,r,s}})\mid_\infty - 2\delta \mid \\
=& \mid\rho(P',\sup{\mathbb{U}_{P,r,s}})-\rho(P',\inf{\mathbb{U}_{P,r,s}})\mid_\infty - 2\delta 
\end{align*}
Since $\sup{\mathbb{U}_{P,r,s}} \notin \mathbb{U}_{P',r,s}$ and $, \inf{\mathbb{U}_{P,r,s}}\notin \mathbb{U}_{P',r,s}$, so we have $$\sup{\mathbb{U}_{P,r,s}}  - \inf{\mathbb{U}_{P,r,s}}  \geq \sup{\mathbb{U}_{P',r,s}}  - \inf{\mathbb{U}_{P',r,s}} $$
Based on the monotonicity of $\rho$ in $l_\infty$ norm given $P$, we have
$$ \mid\rho(P',\sup{\mathbb{U}_{P,r,s}})-\rho(P',\inf{\mathbb{U}_{P,r,s}})\mid_\infty \geq  \mid\rho(P',\sup{\mathbb{U}_{P',r,s}})-\rho(P',\inf{\mathbb{U}_{P',r,s}})\mid_\infty$$
Combining these inequities above, for any $u \in \mathcal{S}$,
\begin{align*}
    \centering
\mid\rho(P, u+\Delta_{OF}^{r,s}) - \rho(P, u)\mid_\infty &\leq \max_{\alpha\in\mathcal{S}}|\rho(P,\alpha+\Delta_{OF}^{r,s}) - \rho(P, \alpha)|_\infty \\
&\leq \min_{P_1\in\mathbb{P}_1}( \mid\rho(P_1,\sup{\mathbb{U}_{P_1,r,s}})-\rho(P_1,\inf{\mathbb{U}_{P_1,r,s}})\mid_\infty-2\delta)\\
&\leq\mid\rho(P',\sup{\mathbb{U}_{P',r,s}})-\rho(P',\inf{\mathbb{U}_{P',r,s}})\mid_\infty -2\delta\\
&\leq \mid\rho(P',\sup{\mathbb{U}_{P,r,s}})-\rho(P',\inf{\mathbb{U}_{P,r,s}})\mid_\infty - 2\delta \\
&\leq \mid\rho(P,\sup{\mathbb{U}_{P,r,s}})-\rho(P,\inf{\mathbb{U}_{P,r,s}})\mid_\infty
\end{align*}
Based on the monotonicity of $\rho$ in $l_\infty$ norm given $P$, let $u=\inf{\mathbb{U}_{P,r,s}}$, we have $\Delta_{OF}^{r,s} \leq \sup{\mathbb{U}_{P,r,s}} - \inf{\mathbb{U}_{P,r,s}}$ for any $P$ and $(r,s)$.
Therefore, for pixel $(r,s)$,
\begin{align}
    \Delta_{OF}^{r,s} \leq  \min_{P\in \mathbb{P}}\{\sup{\mathbb{U}_{P,r,s}} - \inf{\mathbb{U}_{P,r,s}}\} \notag
\end{align}
which concludes the proof.
\end{proof}

\subsection{Proof on Theorem \ref{theorem:diff_res_appro_one}: Certification with approximated partitions from one-frame point cloud}
\begin{theorem}[Certification with approximated partitions from one-frame point cloud, \textbf{restated} of Theorem \ref{theorem:diff_res_appro_one}]
    \label{theorem:diff_res_appro_one_re}
For the  projection from one-frame 3D point cloud $V_1\in\mathcal{V}_1: \mathbb{P}_1\times[0, 1]^K$ and unknown entire point cloud  $V\in\mathcal{V}: \mathbb{P}\times[0, 1]^K$ with $\delta$-convexity, 
let $\phi$ be the one-axis rotation or translation with parameters in $\mathcal{S}\subseteq\mathcal{Z}$ with  the  monotonic projection function $\rho$ in $l_\infty$ norm  over $\mathcal{S}$, we have uniformly partitioned $\{\alpha_i\}_{i=1}^N\subseteq\mathcal{S}$ under interval $\Delta_\alpha$ with  finite constant $C_\delta\geq0$ where
\begin{align}
    \Delta_\alpha \leq  \min_{r,s}\Delta_{OF}^{r,s} & =  \min_{r,s}\min \limits_{P_1\in\mathbb{P}_1}\frac{1}{ L_{P_1} + C_\delta}  \min_{P_1\in\mathbb{P}_1}( \mid\rho(P_1,\sup{\mathbb{U}_{P_1,r,s}})-\rho(P_1,\inf{\mathbb{U}_{P_1,r,s}})\mid_\infty-2\delta) \notag\\
                 &L_{P_1} = \max \{\max_{\alpha\in\mathcal{S}}|\frac{\partial\rho_1(P_1,\alpha)}{\partial\alpha}|, \max_{\alpha\in\mathcal{S}}|\frac{\partial\rho_2(P_1,\alpha)}{\partial\alpha}|\}
\end{align}
Let $y_A, y_B\in\mathcal{Y}$, $\varepsilon\sim\mathcal{N}(0, \sigma^2 \mathbb{I}_d)$ and suppose that for any $i$, the $\varepsilon$-smoothed classifier defined by $q(y\mid x;\varepsilon):=\mathbb{E}(p(y\mid x+\varepsilon))$ has class probabilities that satisfy  with top-2 classes $y_A,y_B$ as
                 $q(y_A\mid \phi(x, \alpha_i); \varepsilon) \geq p_A^{(i)} \geq p_B^{(i)} \geq \max_{y\neq y_A}q(y\mid \phi(x, \alpha_i); \varepsilon)$,

            then it is guaranteed that $\forall\alpha\in\mathcal{S}\colon y_A = \argmax_y q(y\mid \phi(x, \alpha);\varepsilon)$ if 
            $$ \max_{1\leq i\leq N-1}\sqrt{\frac{1}{2} \sum_{k,r,s}(\phi(x, \alpha_i)_{krs}-\phi(x, \alpha_{i+1})_{krs})^2}   < \frac{\sigma}{2}\min_{1\leq i \leq N}\left(\Phi^{-1}\left(p_A^{(i)}\right)-\Phi^{-1}\left(p_B^{(i)}\right)\right).
               $$
\end{theorem}
\begin{proof}
According to Theorem \ref{lemma: diff_res}, we need to upper bound  $\max_{\alpha\in\mathcal{S}}\min_{1\leq i \leq N}\left\|\phi(x, \alpha) - \phi(x, \alpha_i)\right\|_2$ with uniform partitions under interval of $ \Delta_\alpha = \alpha_{i+1} - \alpha_{i}$. 
\begin{align}
    \centering
    &\max_{\alpha\in\mathcal{S}}\min_{1\leq i \leq N}\left\|\phi(x, \alpha) - \phi(x, \alpha_i)\right\|_2 \notag\\\leq & \max_{1\leq i\leq N-1}\sqrt{\max_{\alpha_i\leq \alpha \leq \alpha_{i+1}}\min\{(\phi(x, \alpha)-\phi(x, \alpha_i))^2, (\phi(x, \alpha)-\phi(x, \alpha_{i+1}))^2\} }\notag \\
    = & \max_{1\leq i\leq N-1}\sqrt{\max_{\alpha_i\leq \alpha \leq \alpha_{i+1}}\min\{\sum_{k,r,s}(\phi(x, \alpha)_{krs}-\phi(x, \alpha_i)_{krs})^2,  \sum_{k,r,s}(\phi(x, \alpha)_{krs}-\phi(x, \alpha_{i+1})_{krs})^2\}} \notag \\
    \leq & \max_{1\leq i\leq N-1}\sqrt{\max_{\alpha_i\leq \alpha \leq \alpha_{i+1}}\frac{1}{2}\{\sum_{k,r,s}(\phi(x, \alpha)_{krs}-\phi(x, \alpha_i)_{krs})^2+ \sum_{k,r,s}(\phi(x, \alpha)_{krs}-\phi(x, \alpha_{i+1})_{krs})^2\}} \notag \\
    = & \max_{1\leq i\leq N-1}\sqrt{\max_{\alpha_i\leq \alpha \leq \alpha_{i+1}}\frac{1}{2}\sum_{k,r,s}[(\phi(x, \alpha)_{krs}-\phi(x, \alpha_i)_{krs})^2+ (\phi(x, \alpha)_{krs}-\phi(x, \alpha_{i+1})_{krs})^2]} \notag 
\end{align}
According to Lemma \ref{lemma:find_Delta_appro_one_re}, 
for any $\alpha_i\leq \alpha\leq \alpha_{i+1}$ on the pixel $(r,s)$, \begin{align}
    \Delta_\alpha \leq  \min_{r,s} \Delta_{OF}^{r,s} &\leq \min_{r,s} \min_{P\in \mathbb{P}}\{\sup{\mathbb{U}_{P,r,s}} - \inf{\mathbb{U}_{P,r,s}}\}  \notag \\ & \leq \min_{P\in \mathbb{P}}\{\sup{\mathbb{U}_{P,r,s}} - \inf{\mathbb{U}_{P,r,s}}\}  \notag
\end{align}
Based on Lemma \ref{lemma:find_exact_Delta_re}, we have
$$O(V, \alpha)_{r,s} \in\{O(V, \alpha_i)_{r,s},O(V, \alpha_{i+1})_{r,s}\}$$
i.e., either $O(V, \alpha)_{r,s}=O(V, \alpha_i)_{r,s}$ or $O(V, \alpha)_{r,s}=O(V, \alpha_{i+1})_{r,s}$ holds. With $O(V, \alpha) = \phi(x, \alpha)$ for any $\alpha\in\mathcal{S}$,  for any $k, r, s$, we have
$$(\phi(x, \alpha)_{krs}-\phi(x, \alpha_i)_{krs})^2+ (\phi(x, \alpha)_{krs}-\phi(x, \alpha_{i+1})_{krs})^2 = (\phi(x, \alpha_i)_{krs}-\phi(x, \alpha_{i+1})_{krs})^2$$
Therefore,
\begin{align}
    \centering
    &\max_{\alpha\in\mathcal{S}}\min_{1\leq i \leq N}\left\|\phi(x, \alpha) - \phi(x, \alpha_i)\right\|_2 \notag\\   
    \leq & \max_{1\leq i\leq N-1}\sqrt{\max_{\alpha_i\leq \alpha \leq \alpha_{i+1}}\frac{1}{2}\sum_{k,r,s}[(\phi(x, \alpha)_{krs}-\phi(x, \alpha_i)_{krs})^2+ (\phi(x, \alpha)_{krs}-\phi(x, \alpha_{i+1})_{krs})^2]} \notag \\
    =&\max_{1\leq i\leq N-1}\sqrt{\frac{1}{2} \sum_{k,r,s}(\phi(x, \alpha_i)_{krs}-\phi(x, \alpha_{i+1})_{krs})^2} 
    \notag
\end{align}
So if \begin{equation}
                \begin{gathered} \max_{1\leq i\leq N-1}\sqrt{\frac{1}{2} \sum_{k,r,s}(\phi(x, \alpha_i)_{krs}-\phi(x, \alpha_{i+1})_{krs})^2}   < \frac{\sigma}{2}\min_{1\leq i \leq N}\left(\Phi^{-1}\left(p_A^{(i)}\right)-\Phi^{-1}\left(p_B^{(i)}\right)\right).
                \end{gathered}
            \end{equation}
 then it holds that
\begin{equation}
                \label{eq:general_Ms_re}
                \begin{gathered}
                    M_{\mathcal{S}}:= \max_{\alpha\in\mathcal{S}}\min_{1\leq i \leq N}\left\|\phi(x, \alpha) - \phi(x, \alpha_i)\right\|_2< R:=\frac{\sigma}{2}\min_{1\leq i \leq N}\left(\Phi^{-1}\left(p_A^{(i)}\right)-\Phi^{-1}\left(p_B^{(i)}\right)\right).
                \end{gathered}
            \end{equation}
which concludes the proof based on Theorem \ref{lemma: diff_res}.
\end{proof}

\section{More Experiment Details and Results}
\label{sec:E}
In this section, we present more details of the experiments, including experimental setup, model training, and more results and analysis. 
\subsection{Experimental Setup}
\label{sec:E_1}
\textbf{Model training.} For the model training, we first mix  the training sets of all motion types, $T_z, T_x, T_y, R_z, R_x, R_y$, generating the whole training set for data augmentation. We adopt the ImageNet \cite{deng2009imagenet} pre-trained models with architectures of ResNet101 and ResNet50 \cite{he2016deep} from torch vision and fine-tune them for 100 epochs using the six mixed motion augmented training data \cite{hu2022robustness} with a learning rate of 0.001  and consistency regularization \cite{jeong2020consistency, hu2022robustness}.  For the base model with diffusion-based denoiser \cite{carlini2022certified}, we adopt the pre-trained unconditional 256x256 diffusion model \cite{dhariwal2021diffusion} as the frozen denoiser  and only fine-tune the base classifier for 1 epoch to address the domain shift. To fairly compare the certification performance, we keep the base classifiers the same for the baseline and ours. Please note that diffusion-based denoiser \cite{carlini2022certified} does not apply to CMS baseline because the denoiser is designed for pixel-wise Gaussian denoising as the pre-trained model and can only apply to pixel-wise smoothing.



\textbf{Details of the required projection frames.}
Since image capturing from the camera is the most non-efficient and time-consuming part of the certification against camera motion, we adopt the \textit{number of required projected frames} in the certification as the fair hardware-independent metric to compare the certification efficiency. Theoretically, the number of required projection frames for the uniform partitions can be calculated based on Lemma \ref{lemma:find_exact_Delta}, \ref{lemma:find_Delta_appro} and \ref{lemma:find_Delta_appro_one} for \textit{PwS}, \textit{PwS-L}, \textit{PwS-OF} in the implementation of MetaRoom dataset. Empirically, it can be observed that the certification performance will converge as the partition number  increases, as shown in Section \ref{sec:abla}. Therefore, to avoid redundant partitions in the implementation, we adopt a  quantile of over 99\% for the minimal fully-covered partition interval across all the pixels as the required offline projection frames and take the average over all images.
%
%
%

\begin{table}[htb]
\caption{Required projection frames as partitions and corresponding certified accuracy for different axes of rotation and translation, with ResNet-50 and $\sigma=0.5$}
\centering
\begin{tabular}{cccc}
\toprule
\begin{tabular}[c]{@{}c@{}}Num. of  Required  Projected \\ Frames / Certified Accuracy\end{tabular} 
       &PwS          & PwS-L        & PwS-OF         \\\midrule
x-axis translation $T_x$, 5mm radius                 & 3.1k / 0.192  & 5.0k / 0.192 & 6.1k / 0.192\\
y-axis translation $T_y$, 5mm radius                 & 3.2k / 0.183 & 4.8k / 0.183 & 6.0k / 0.183 \\
x-axis rotation $R_x$, 0.25$^\circ$ radius                  & 3.8k / 0.183 & 5.3k / 0.192 & 6.5k / 0.192 \\
z-axis rotation $R_z$,  0.7$^\circ$ radius                  & 3.2k / 0.133 & 4.9k / 0.133 & 5.8k / 0.133 \\\bottomrule
\end{tabular}
\label{tab:all_frames_acc}
\end{table}

\begin{table}[htb]
\caption{Certified accuracy with different smoothing variance $\sigma^2$ in all projections under ResNet-101}
\centering
\begin{tabular}{ccccccc}
\toprule
Projection and radius & $T_z$, 10mm & $T_x$, 5mm & $T_y$, 5mm & $R_z$, 0.7$^\circ$ & $R_x$,  0.25$^\circ$ & $R_y$, 0.25$^\circ$ \\ \midrule
 $\sigma=0.25$    &  0.175  &   0.0      &   0.075      & 0.0        &      0.058     &  0.025        \\
$\sigma=0.5$    & 0.150   &   0.167      &    0.133     & 0.117        &   0.142        &    0.125     \\
$\sigma=0.75$     & 0.140    &  0.100       &  0.133       &     0.075    &     0.117      &  0.100  \\\bottomrule  
\end{tabular}
\label{tab:all_variance_101}
\end{table}

\begin{table}[htb]
\caption{Certified accuracy under different smoothing variances and different radii}
\centering
\begin{tabular}{ccccc}
\toprule Radii   along z-axis translation, $T_z$                  & \multicolumn{2}{c}{10mm}                           & \multicolumn{2}{c}{100mm}                          \\
\multicolumn{1}{c}{Smoothing variance $\sigma^2$, PwS } & \multicolumn{1}{c}{$\sigma=0.25$} & \multicolumn{1}{c}{$\sigma=0.5$} & \multicolumn{1}{c}{$\sigma=0.25$} & \multicolumn{1}{c}{$\sigma=0.5$} \\\midrule
ResNet50                    &     0.198               & 0.223                    &       0.0             & 0.142                    \\
ResNet101                   &       0.175             & 0.150                    &      0.0              & 0.108   \\\bottomrule                  
\end{tabular}
\label{tab:all_complexity}
\end{table}

\begin{table}[htb]
\caption{Certified accuracy with different numbers of  partitioned images in pixel-wise smoothing under different radii}
\centering
\begin{tabular}{cccccccc}
\toprule
\begin{tabular}[c]{@{}c@{}}Number of Partitions\\ Resnet-50,  z-axis translation $T_z$\end{tabular} & 1000  & 2000  & 3000  & 4000  & 5000  & 6000  & 7000  \\ \midrule
10mm radius                                                                          & 0.183 & 0.192 & 0.217 & 0.223 & 0.223 & 0.223 & 0.231 \\
20mm radius                                                                        & 0.167 & 0.175 & 0.183 & 0.192 & 0.192 & 0.192 & 0.200 \\ \bottomrule
\end{tabular}
\label{tab:all_partition}
\end{table}

\textbf{Metric of average certification time per image.} In addition to the \textit{number of required projected frames}, we also report the average certification time per image as an efficiency metric. We conduct all the experiments on the same machine with NVIDIA A6000 and 512G RAM so that it is fair to compare the time consumption of different methods. To calculate the average time per image, we collect the time duration for every test sample, starting from reading the camera pose and ending with deciding whether the certification condition holds or not using Monte Carlo based algorithm \cite{cohen2019certified}. We then find the mean of all the time duration and report it as the average certification time per image.

\textbf{Empirical robust accuracy and benign accuracy}
We further compare the empirical robust accuracy and benign accuracy of the smoothed classifier added with pixel-wise Gaussian smoothing distribution. Following the metric in previous work \cite{li2021tss, hu2022robustness}, we adopt 100-perturbed worst-case empirical robust accuracy, which is calculated as follows. By making 100 uniform perturbations with the attack radius in the camera motion space, we feed every motion-perturbed image into the smoothed classifier; if any of these 100 perturbed images are wrongly classified, we report the smoothed classifier is not robust at  the test sample. The empirical robust accuracy is calculated as the ratio of  robust test samples among all the test samples. The benign accuracy is the correctly-classified ratio of the test samples without any camera motion perturbation. The certified accuracy is at the same certified radius as the perturbation radius.

\begin{table}[htb]
\caption{Certification time per image and perception ratio comparison with baseline CMS}
\centering
\begin{tabular}{ccccc}
\toprule
Certification time (s / image) & CMS \cite{hu2022robustness}            & PwS            & PwS-L          & PwS-OF         \\\midrule
$T_z$, 10mm radius                & 2085.8 / 100\% & 172.9 / 8.3\%  & 279.8 / 13.2\% & 479.5 / 23.0\% \\
$R_y$, 0.25 radius                & 2035.4 / 100\% & 236.4 / 11.6\% & 398.7 / 19.6\% & 608.5 / 29.9\% \\ \bottomrule
\end{tabular}
\label{tab:time_compare}
\end{table}
\begin{table}[h]
\centering
\caption{Comparison of certified accuracy and empirical robust accuracy.  The pixel-wise smoothed classifier is based on ResNet-50 with $\sigma=0.75$}
\label{tab:erm}
\begin{tabular}{cccc}
\toprule
\begin{tabular}[c]{@{}c@{}}Perturbation radii \\ for $T_z$ (mm)\end{tabular} & \begin{tabular}[c]{@{}c@{}}Certified \\ Accuracy\end{tabular} & \begin{tabular}[c]{@{}c@{}}Empirical\\ Robust Accuracy\end{tabular} & \begin{tabular}[c]{@{}c@{}}Benign\\  Accuracy\end{tabular}\\ \midrule
{[}-10, 10{]}         &  0.140                                                             &  0.283             &                                       \multirow{3}{*}{0.308}           \\ 
{[}-20, 20{]}         &     0.117                                                          &   0.275                                                       &       \\ 
{[}-100, 100{]}       &   0.091                                                            &   0.258           &                                                   \\ \bottomrule
\begin{tabular}[c]{@{}c@{}}Perturbation radii \\  for $R_y$ ($^\circ$) \end{tabular}
 & \begin{tabular}[c]{@{}c@{}}Certified \\ Accuracy\end{tabular} & \begin{tabular}[c]{@{}c@{}}Empirical\\ Robust Accuracy\end{tabular}& \begin{tabular}[c]{@{}c@{}}Benign\\  Accuracy\end{tabular} \\ \midrule
{[}-0.25, 0.25{]}         &  0.108                                                            &  0.275                                                     &                                       \multirow{3}{*}{0.316}          \\ 
{[}-0.5, 0.5{]}         &  0.092                                                             &    0.258       &                                                    \\ 
{[}-2.5, 2.5{]}       &   0.058                                                            &   0.117             &                                                 \\ \bottomrule
\end{tabular}
\end{table}

\subsection{More Experimental Results}
\label{sec:E_2}
\textbf{Projected frames required and certified accuracy for more axes projection.}
From Table \ref{tab:all_frames_acc}, we can see that for each translation or rotation, the number of required projected frames of \textit{PwS} is less than \textit{PwS-L} and \textit{PwS-OF}, which is consistent with the theoretical analysis of Lemma \ref{lemma:find_exact_Delta}, \ref{lemma:find_Delta_appro} and \ref{lemma:find_Delta_appro_one}. Also, the certified accuracy is very close due to the convergence and saturation  of certification performance regarding partition number.

\textbf{Influence of the variance of the smoothing distribution on ResNet-101.}
Similar to Table \ref{tab:abla_variance},  the certified accuracy with $\sigma=0.5$ is  no better than  that with $\sigma=0.75$ due to  the accuracy/robustness trade-off \cite{cohen2019certified} for all translation and rotation axes in Table \ref{tab:all_variance_101}. Different from Table \ref{tab:abla_variance}, the  performance with $\sigma=0.25$ is better for z-axis translation than $\sigma=0.5$ while much poorer for other axes, which shows that the more complex model is less sensitive to z-axis translation to cover the projection errors in the certification condition.

\textbf{Performance under different model complexity with smaller variance.}
Different from Table \ref{tab:abla_complexity}, under smaller variance of pixel-wise smoothing in Table \ref{tab:all_complexity}, larger model complexity  can be more certifiably robust with lower certified accuracy, showing less influence of robust overfitting with smaller variance.

\textbf{Performance with different partitions along z-axis translation.}
We report the influence of  partition numbers in empirical experiments under different models along z-axis translation in Table \ref{tab:all_partition}. Similar to Table \ref{tab:abla_partition}, the certification accuracy will be saturated  as the partition number increases, which covers the required projected frames in Table \ref{tab:num_of_frames}.

\textbf{Comparison of Certification Time Efficiency.}
We compare the time efficiency for the baseline CMS \cite{hu2022robustness} and ours in Table \ref{tab:time_compare}. Note that all the baselines of CMS are with 10k Monte Carlo sampling as default \cite{hu2022robustness} for the comparison. We can see that our method has significantly less wall-clock time compared to the baseline in terms of certification time per image, due to requiring fewer projected frames offline.

\textbf{Comparison with Empirical Robust Accuracy and Benign Accuracy.}
As shown in Table \ref{tab:erm}, we compare the certified accuracy, empirical robust accuracy, and benign accuracy for smoothed classifier of ResNet50 with $\sigma=0.75$. It can be seen that the certified accuracy is lower than the empirical robustness accuracy, which is consistent with the claim in the previous work  \cite{li2021tss, hu2022robustness}. The empirical robust accuracy is lower than the benign accuracy, showing that even  the pixel-wise smoothed classifier is not robust against the camera motion perturbation and needs certification to provable guarantee the robustness.
Please note that we adopt the smoothed classifier,  which requires the input image to be added pixel-wise zero-mean Gaussian noise as smoothing. This is the reason why the empirical robust accuracy and benign accuracy are  quite low compared to that   in the previous work \cite{hu2022robustness}, owing to the accuracy/robustness trade-off of smoothing distribution \cite{cohen2019certified}.


\end{document}